\documentclass{article}
\PassOptionsToPackage{numbers}{natbib}

 \usepackage[main, final]{neurips_2026}

\usepackage[utf8]{inputenc} 
\usepackage[T1]{fontenc}    
\usepackage{hyperref}       
\usepackage{url}            
\usepackage{booktabs}       
\usepackage{amsfonts}       
\usepackage{nicefrac}       
\usepackage{microtype}      
\usepackage{xcolor}         

\usepackage{microtype}
\usepackage{graphicx}
\usepackage{subcaption}
\usepackage{booktabs} 
\usepackage[inline]{enumitem}

\usepackage{hyperref}
\usepackage{url}
\usepackage{caption}
\usepackage{multirow} 
\usepackage{siunitx}
\usepackage{setspace}
\usepackage{mathtools}

\usepackage[ruled,vlined]{algorithm2e}
\usepackage{xcolor}
\usepackage{wrapfig}
\usepackage{framed}
\usepackage{xcolor}
\usepackage{tcolorbox}
\tcbuselibrary{skins}
\usepackage{pifont}
\usepackage{tikz}
\usetikzlibrary{calc}

\usepackage{amsfonts}

\usepackage{hyperref}

\newtcolorbox{keyfindingbox_sl}{
    enhanced,
    colback=gray!10,      
    colframe=gray!70,   
    arc=3mm,                  
    boxrule=1pt,
    drop shadow={white}, 
    boxsep=5pt,               
    left=6pt, right=6pt, top=4pt, bottom=4pt 
}

\newtcolorbox{keyfindingbox}{
    enhanced,
    colback=gray!10,      
    colframe=gray!70,   
    arc=3mm,                  
    boxrule=1pt,
    drop shadow={white}, 
    boxsep=5pt,               
    left=40pt, right=40pt, top=16pt, bottom=16pt 
}

\newenvironment{denseitemize}{
	\begin{itemize}[topsep=2pt, partopsep=0pt, leftmargin=1.5em]
		\setlength{\itemsep}{2pt}
		\setlength{\parskip}{0pt}
		\setlength{\parsep}{0pt}
	}{\end{itemize}}

\definecolor{mars-url-teal}{HTML}{008C8C}
\definecolor{git-red-bg}{rgb}{1.0,0.8,0.8}
\definecolor{git-green-bg}{rgb}{0.8,1.0,0.8}
\newlength{\algofullwidth}

\newcommand{\indenteddeletedblock}[1]{%
  \noindent\colorbox{git-red-bg}{%
    \parbox{\dimexpr\algofullwidth-2.6em-2\fboxsep\relax}{#1}}%
}
\newcommand{\indentedaddedblock}[1]{%
  \noindent\colorbox{git-green-bg}{%
    \parbox{\dimexpr\algofullwidth-2.6em-2\fboxsep\relax}{#1}}%
}

\usepackage{amsmath}
\usepackage{amssymb}
\usepackage{mathtools}
\usepackage{amsthm}

\usepackage[capitalize,noabbrev]{cleveref}

\theoremstyle{plain}

\theoremstyle{definition}

\theoremstyle{remark}

\title{MARS: Harmonizing Multimodal Convergence via Adaptive Rank Search}

\author{%
  \textbf{Minkyoung Cho\textsuperscript{1}\quad
  Insu Jang\textsuperscript{1}\quad
  Shuowei Jin\textsuperscript{1}\quad
  Zesen Zhao\textsuperscript{1}\quad
  Adityan Jothi\textsuperscript{2}}\\[0.35em]
  \textbf{Ethem F. Can\textsuperscript{2}\quad
  Min-Hung Chen\textsuperscript{2,$\dagger$}\quad
  Z. Morley Mao\textsuperscript{1,$\dagger$}}\\[0.65em]
  \textsuperscript{1}University of Michigan
  \qquad
  \textsuperscript{2}NVIDIA\\[0.5em]
  \textnormal{\href{https://minkyoungcho.github.io/mars/}{%
    \textcolor{mars-url-teal}{https://minkyoungcho.github.io/mars/}}}
}

\newcommand\blfootnote[1]{%
  \begingroup
  \renewcommand\thefootnote{}\footnotetext{#1}%
  \endgroup
}

\begin{document}

\maketitle
\blfootnote{\textsuperscript{$\dagger$}Equal advising.\quad Correspondence: Minkyoung Cho (\href{mailto:minkycho@umich.edu}{minkycho@umich.edu})}

\begin{abstract}
  \looseness=-1
  Fine-tuning Vision Language Models (VLMs) with parameter-efficient methods like Low-Rank Adaptation (LoRA) is crucial for task adaptation. However, imbalanced training dynamics across modalities often lead to suboptimal accuracy due to negative interference, a challenge typically addressed with inefficient heuristic methods such as manually tuning separate learning rates. 
  To overcome this, we introduce \textbf{MARS} (\textbf{M}ultimodal \textbf{A}daptive \textbf{R}ank \textbf{S}earch), an approach to discover optimal rank pairs that balance training dynamics while maximizing performance.
  Our key innovation, a proposed framework of dual scaling laws, enables this search:~one law models module-specific convergence time to prune the search space to candidates with aligned dynamics, while the other predicts final task performance to select the optimal pair from the pruned set.
  By re-purposing the LoRA rank as a controller for modality-specific convergence speed, MARS outperforms baseline methods and provides a robust, automated strategy for optimizing VLM fine-tuning. 
  \end{abstract}
  
  \section{Introduction}
  Vision Language Models (VLMs) are built from three key components: a vision encoder (VE) that extracts visual representations, a projector that bridges visual and language spaces, and a large language model (LLM) backbone that reasons over the combined inputs. While early approaches kept the VE frozen and adapted only the projector or LLM, it has become increasingly clear that deeply integrated multimodal understanding requires fine-tuning all components jointly~\cite{kim2024openvla}. This has established comprehensive end-to-end fine-tuning as the dominant paradigm for state-of-the-art VLMs~\cite{zhang2024mm1,zhai2024berkerlyft,zanella2024loraclip,chen2025janus}. However, given the immense scale of these models, full fine-tuning is computationally prohibitive, making parameter-efficient methods indispensable. Low-rank adaptation (LoRA)~\cite{hu2022lora}, which approximates weight updates with low-rank matrices, has emerged as the standard approach for enabling this comprehensive adaptation at tractable cost.
  
\begin{figure}[t]
    \centering
    \begin{subfigure}{0.49\linewidth}
        \centering
        \includegraphics[width=0.49\linewidth]{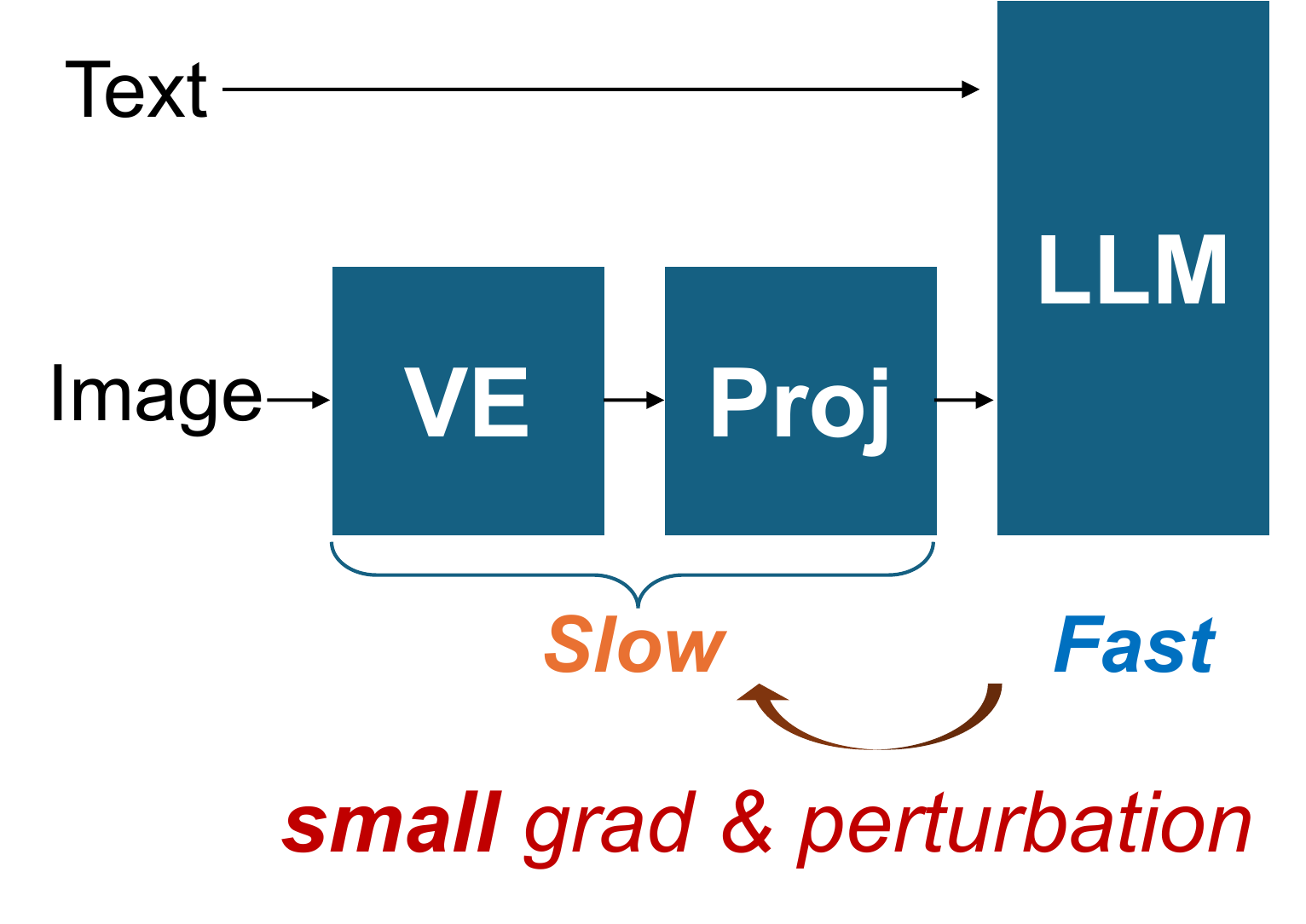}
        \hfill
        \includegraphics[width=0.49\linewidth]{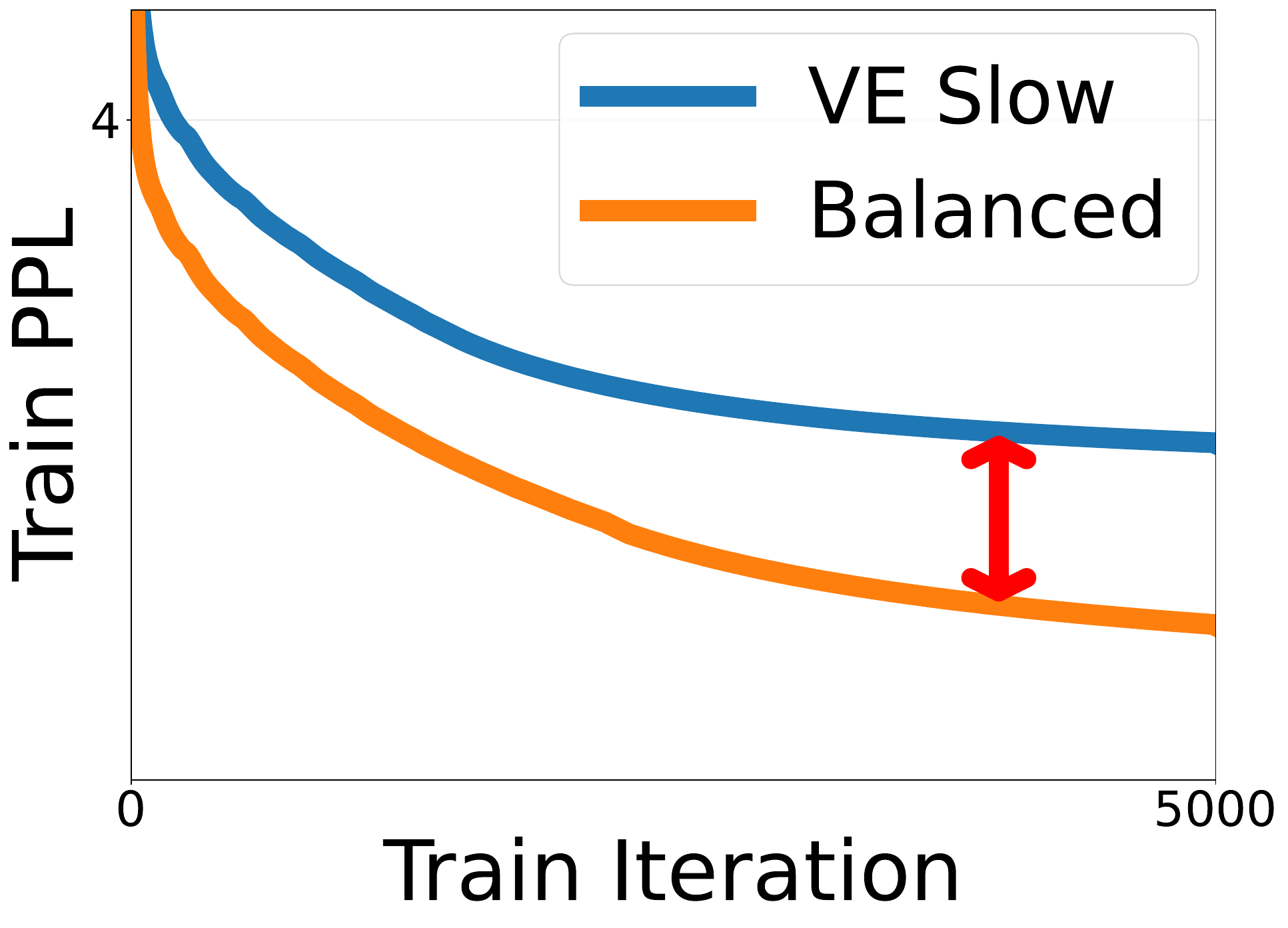}
        \caption{VE under-adapted causes a performance bottleneck.}
        \label{fig:motivation_me_group}
    \end{subfigure}%
    \hfill
    \begin{subfigure}{0.49\linewidth}
        \centering
        \includegraphics[width=0.49\linewidth]{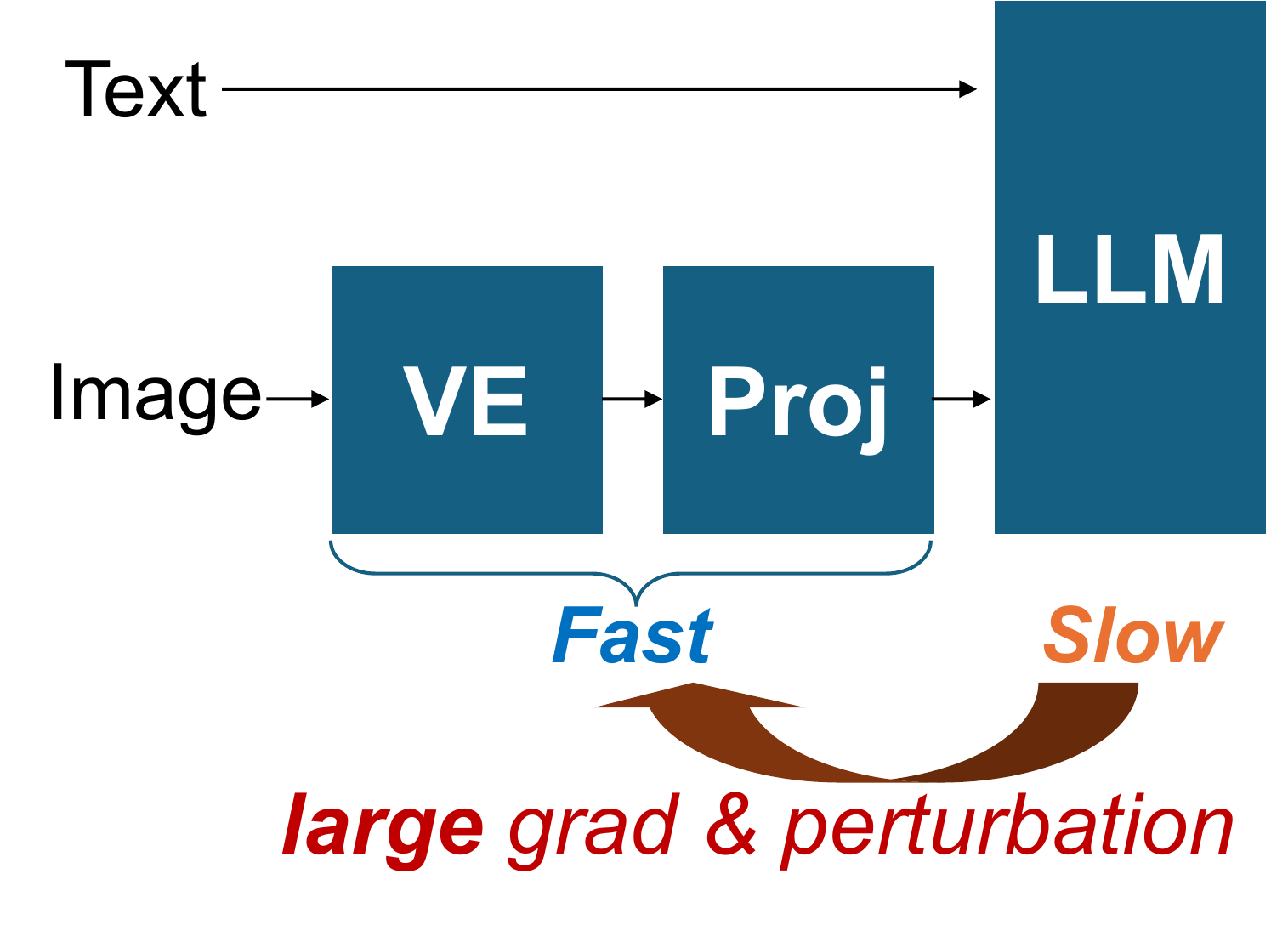}
        \hfill
        \includegraphics[width=0.49\linewidth]{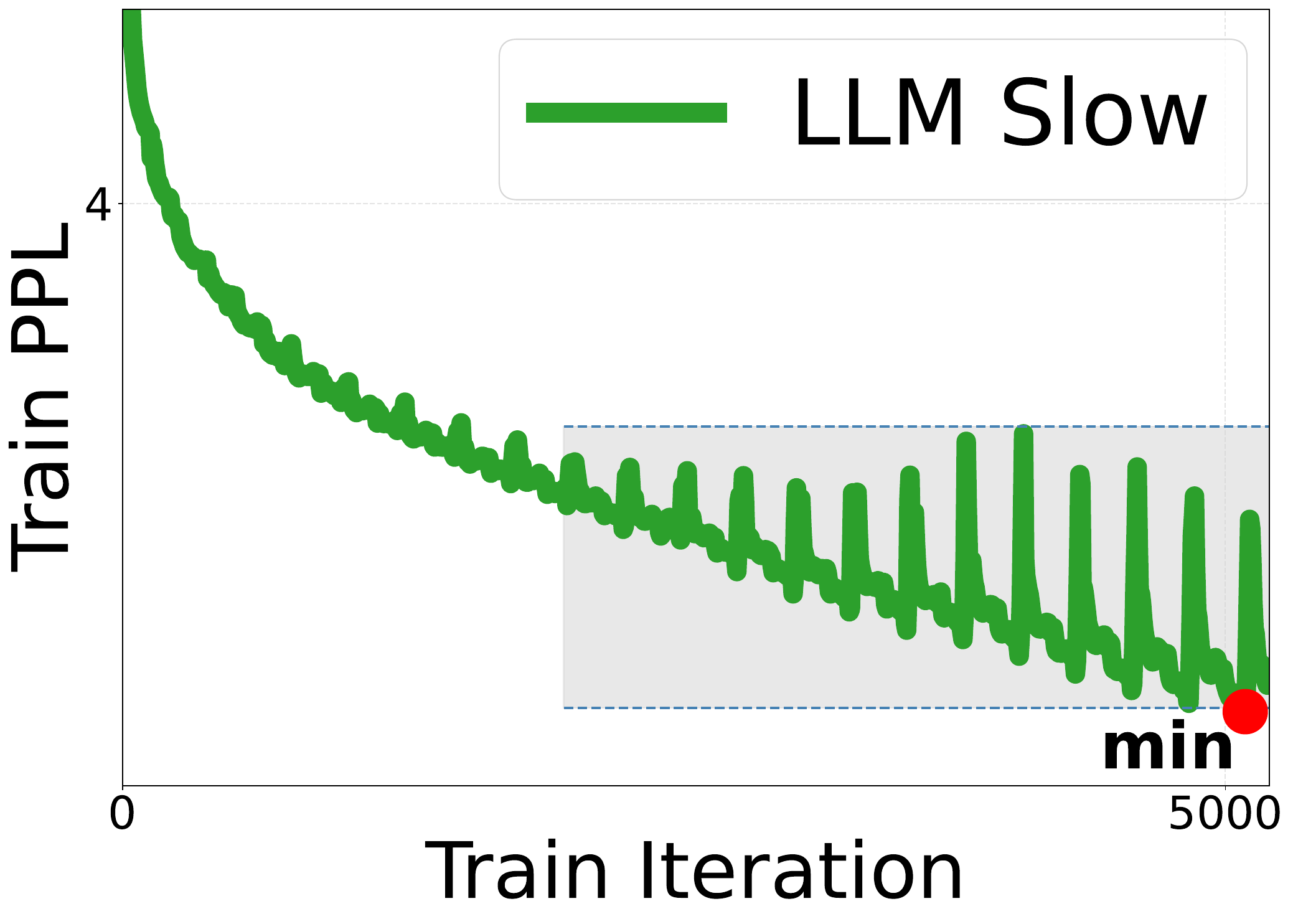}
        \caption{LLM under-adapted causes training oscillations.}
        \label{fig:motivation_llm_group}
    \end{subfigure}
    \caption{\textbf{Motivation: Imbalanced training dynamics lead to suboptimal performance.}
    Conceptual cases of module imbalance (left) and their corresponding empirical results (right).}
    \label{fig:motivation_fig1}
\end{figure}

  \looseness=-1
  In practice, LoRA is typically applied uniformly across the VE and LLM, treating the two modules identically without differentiation~\cite{zhai2024berkerlyft, zanella2024loraclip, schotthofer2024geolora}. However, because the VE and LLM originate from distinct pre-trained unimodal models with different parameter scales, they exhibit fundamentally different learning dynamics and converge at different rates under a shared training configuration. As illustrated in Figure~\ref{fig:motivation_fig1}, this imbalance leads to performance bottlenecks and training oscillations. A natural remedy is to assign differential learning rates to each module, but this is laborious in practice and depends on costly trial-and-error search~\cite{li2024llava-ov, bai2025qwen25, zhang2024mm1}. Moreover, since the learning rate only scales gradient magnitudes, it only addresses the symptom -- speed of convergence -- rather than the root cause. A more fundamental lever is the LoRA rank itself, which directly controls each module's adaptation capacity and acts as a regularizer~\cite{biderman2024loralearnless}. Assigning differential ranks therefore offers a principled way to harmonize the training dynamics of the VE and LLM.
  However, identifying an optimal rank pair is challenging due to an inherent \textit{two-fold disparity}: (1) a disparity in learning capacity, stemming from their differing parameter scales, and (2) a disparity in the required learning budget, as each module originates from a distinct pre-trained unimodal model with its own domain gap to the downstream task. Therefore, the core challenge lies in quantifying these disparities to align the convergence dynamics of all modules.

  \looseness=-1
  To address this, we introduce \textbf{MARS} (\textbf{M}ultimodal \textbf{A}daptive \textbf{R}ank \textbf{S}earch), an automated algorithm that assigns module-specific LoRA ranks to align the convergence dynamics of the VE and LLM. By searching for the rank pair that harmonizes each module's learning trajectory, MARS eliminates the need for heuristic tuning and systematically resolves the two-fold disparity.
  However, naively searching over all rank pairs is intractable; the combinatorial space is vast, and evaluating each candidate requires a full fine-tuning run. To make MARS efficient, we propose dual scaling laws tailored for VLM fine-tuning~\cite{kaplan2020sc_kaplan,zhang2024sc_llmft,shukor2025sc_mm}: \textbf{Scaling Law-C} (Convergence) captures how each module's convergence rate scales with its LoRA rank, and \textbf{Scaling Law-P} (Performance) captures how final task performance varies with the rank pair. Their free parameters are estimated through a lightweight calibration phase on the target model. Armed with these laws, MARS efficiently identifies the optimal rank pair---retaining only candidates where both modules converge simultaneously via Scaling Law-C, and selecting the highest-performing one via Scaling Law-P---without running a single full fine-tuning job.
  
  Our main contributions are:

  \begin{denseitemize}
      \item We identify and provide evidence that the imbalanced training dynamics in VLM fine-tuning, originating from a two-fold disparity, represent a key source of suboptimal performance. To overcome this, we propose MARS, an automated algorithm that systematically mitigates the imbalance by discovering optimal, modality-specific LoRA rank pairs.
      \item We are the first to propose and validate dual scaling laws for VLM fine-tuning that model module-specific convergence time (Scaling Law-C) and performance (Scaling Law-P), making the rank search feasible.
      \item MARS consistently outperforms fixed/adaptive-rank baselines, learning-rate tuning, and equal-budget search methods across diverse model–benchmark settings, while requiring only lightweight calibration and a single fine-tuning run ($11.5\times$ faster than exhaustive search).
    \end{denseitemize}
  
  \section{Related Work and Motivation}
  \label{sec:related_work}
  \paragraph{Fine-Tuning Strategies for VLMs.}
  State-of-the-art VLMs fine-tune all major components -- vision encoder, projector, and LLM backbone -- using LoRA~\cite{hu2022lora} to keep adaptation tractable at scale~\cite{zhang2024mm1,zhai2024berkerlyft,zanella2024loraclip,chen2025janus,kim2024openvla}. However, prior work uniformly applies the same LoRA rank across all modules, treating them as homogeneous despite their fundamentally different parameter scales and pre-training origins~\cite{liu2023llava,li2024llava-ov,wang2024qwen2,bai2025qwen25}. To partially compensate, practitioners resort to heuristically tuning module-specific learning rates; however, this process demands substantial manual effort and is acutely sensitive to miscalibration. Figure~\ref{fig:motivation_fig1} illustrates the cost of such errors: a much lower VE rate ($lr_{ve} \ll lr_{llm}$) induces a severe performance bottleneck, while a much lower LLM rate ($lr_{llm} \ll lr_{ve}$) destabilizes training entirely. Differential learning rates thus offer only a partial solution; they modulate the speed of gradient updates but leave the fundamental mismatch in adaptation capacity unresolved.

  \paragraph{Scaling Laws.}
  \looseness=-1
  Scaling laws have been established across LLM pre-training~\cite{kaplan2020sc_kaplan,hoffmann2022training}, LLM fine-tuning~\cite{zhang2024sc_llmft}, and multimodal pre-training~\cite{shukor2025sc_mm,he2025mordalautomatedpretrainedmodel}, demonstrating that final task performance scales predictably with compute, data, and model size. However, all of these focus on aggregate, end-to-end performance as the prediction target. No prior work has examined scaling behavior at the level of individual modules, nor studied how convergence dynamics of heterogeneous components, such as the VE and LLM, scale with their LoRA rank during fine-tuning.
  
  \paragraph{Motivation.}
  \looseness=-1
  Adjusting learning rates is limited; hence, adjusting the LoRA rank is a more principled alternative, as the rank directly governs each module’s adaptation capacity and implicitly regularizes fine-tuning~\cite{biderman2024loralearnless}. However, identifying the appropriate rank pair is far from straightforward due to an inherent \textit{two-fold disparity} between the VE and LLM. First, the two modules operate at vastly different parameter scales, giving each a distinct ceiling on adaptation. Second, each originates from a separate pretrained unimodal model with its own domain gap to the downstream task, making the adaptation effort inherently unequal across modules. These asymmetries call for a principled, data-driven method for rank selection, inspired by the predictive power of scaling laws in characterizing model behavior.

\section{Multimodal Adaptive Rank Search (MARS) Methodology}
MARS is a pre-fine-tuning procedure that identifies the optimal modality-specific LoRA rank pair $(r^*_{\text{ve}}, r^*_{\text{llm}})$ to align the convergence dynamics of the VE and LLM before task-specific training begins.
MARS addresses this by introducing dual scaling laws that characterize how rank governs module behavior: \textbf{Scaling Law-C} models the relationship between LoRA rank and each module's convergence speed, while \textbf{Scaling Law-P} models the relationship between the rank pair and final task performance (\S~\ref{sec:dual_scaling_laws}).
The guided search algorithm then uses these laws to efficiently find the optimal pair -- using Scaling Law-C to prune the space to convergence-aligned candidates and Scaling Law-P to select the best-performing one -- without running a single full fine-tuning job, as shown in Algorithm~\ref{alg:mars} (\S~\ref{sec:search}).
The parameters required by the dual scaling laws are empirically estimated through a lightweight calibration phase that precedes the guided search (\S~\ref{sec:calibration}).
  
\noindent
\begin{minipage}[t]{0.60\linewidth}
  \begin{algorithm}[H]
    {\footnotesize
    \setlength{\algofullwidth}{\dimexpr\linewidth-15pt\relax}
    \caption{Comparison of Naive Search and MARS}
    \label{alg:mars}
    \DontPrintSemicolon
    \SetKwInOut{Input}{Input}
    \SetKwInOut{Output}{Output}

    \Input{Candidate ranks $R_{\text{options}}$, target dataset $D_f$}
    \Output{Optimal config $(r^*_{\text{ve}},r^*_{\text{llm}})$}
    $BestPerformance \gets -\infty$\;
    $CandidatePairs \gets \emptyset$\;

    \For{$r_{\text{llm}} \in R_{\text{options}}$}{
      \indenteddeletedblock{
        \# \texttt{-{}-{}-} Naive Search: Iterate over all rank pair combinations; full fine-tune to obtain accuracy. \texttt{-{}-{}-}
        \par\smallskip
        \For{$r_{\text{ve}} \in R_{\text{options}}$}{
          $Perf \gets \textsc{FullFineTuning}(r_{\text{ve}},r_{\text{llm}},D_f)$\;
          \If{$Perf > BestPerformance$}{
            $BestPerformance \gets Perf$\;
            $(r^*_{\text{ve}},r^*_{\text{llm}}) \gets (r_{\text{ve}},r_{\text{llm}})$\;
          }
        }
      }
      \hspace*{-0.6em}
      \indentedaddedblock{
        \# +++ MARS: No inner loop; prune the search space and select the best VE rank for $r_{\text{llm}}$. +++
        \par\smallskip
        $r'_{\text{ve}} \gets \textsc{FindBalancedRank}(r_{\text{llm}})$ \tcp*[r]{SL-C}
        Add $(r'_{\text{ve}},r_{\text{llm}})$ to $CandidatePairs$\;
      }
    }
    \hspace*{0.35em}
    \indentedaddedblock{
    \noindent
      \resizebox{\linewidth}{!}{%
        $(r^*_{\text{ve}},r^*_{\text{llm}}) \gets \textsc{SelectBest}(CandidatePairs)$\hfill\texttt{ // SL-P}}
    }
    \Return $(r^*_{\text{ve}},r^*_{\text{llm}})$\;
    }
  \end{algorithm}
\end{minipage}%
\hfill
\begin{minipage}[t]{0.37\linewidth}

  \subsection{Dual Scaling Laws}
  \label{sec:dual_scaling_laws}
  Dual scaling laws are the predictive foundation that makes rank search feasible. In this section, we elaborate on how they are formulated and validated in the VLM fine-tuning setup.

  \subsubsection{Experimental Setup}
  \label{sec:sc_setup}
  To empirically derive our scaling laws, we designed a controlled experimental setup to isolate and measure pure fine-tuning capability.
   Our experiments are based on the LLaVA-OneVision (OV) architecture~\cite{li2024llava-ov}, but we initialize our models ``from scratch.'' This means we assemble publicly available vision encoder (SigLIP;~\cite{zhai2023siglip}) and LLM (Qwen2;~\cite{wang2024qwen2}) checkpoints with a pre-trained projector, ensuring the model architecture is consistent  while  the in-
\end{minipage}

 itial parameters have no prior exposure to the downstream fine-tuning datasets. 
 This controlled initialization allows us to focus solely on the scaling dynamics of the fine-tuning process itself. To study the impact of parameter disparity, we used two variants: LLaVA-OV-0.5B* (minimal VE-LLM parameter size gap) and LLaVA-OV-7B* (significant gap). 
  
  Our primary fine-tuning dataset was LLaVA-158K~\cite{liu2023llava,li2024llava-ov}, from which we sampled varying subset sizes to study scaling trends. We use validation perplexity as our primary metric to measure fundamental fine-tuning capability (transferability), as it directly aligns with the training objective. Experimental setup details are in Appendix~\ref{app:exp_details}. Convergence is defined by a standard early stopping criterion~\cite{prechelt2002earlystop,pytorchearlystop}: the model is considered converged if validation performance fails to improve for a set number of steps (five in this paper).

    \begin{figure*}[t]
      \centering
      \begin{subfigure}[t]{0.49\textwidth}
          \centering
          \includegraphics[width=\linewidth]{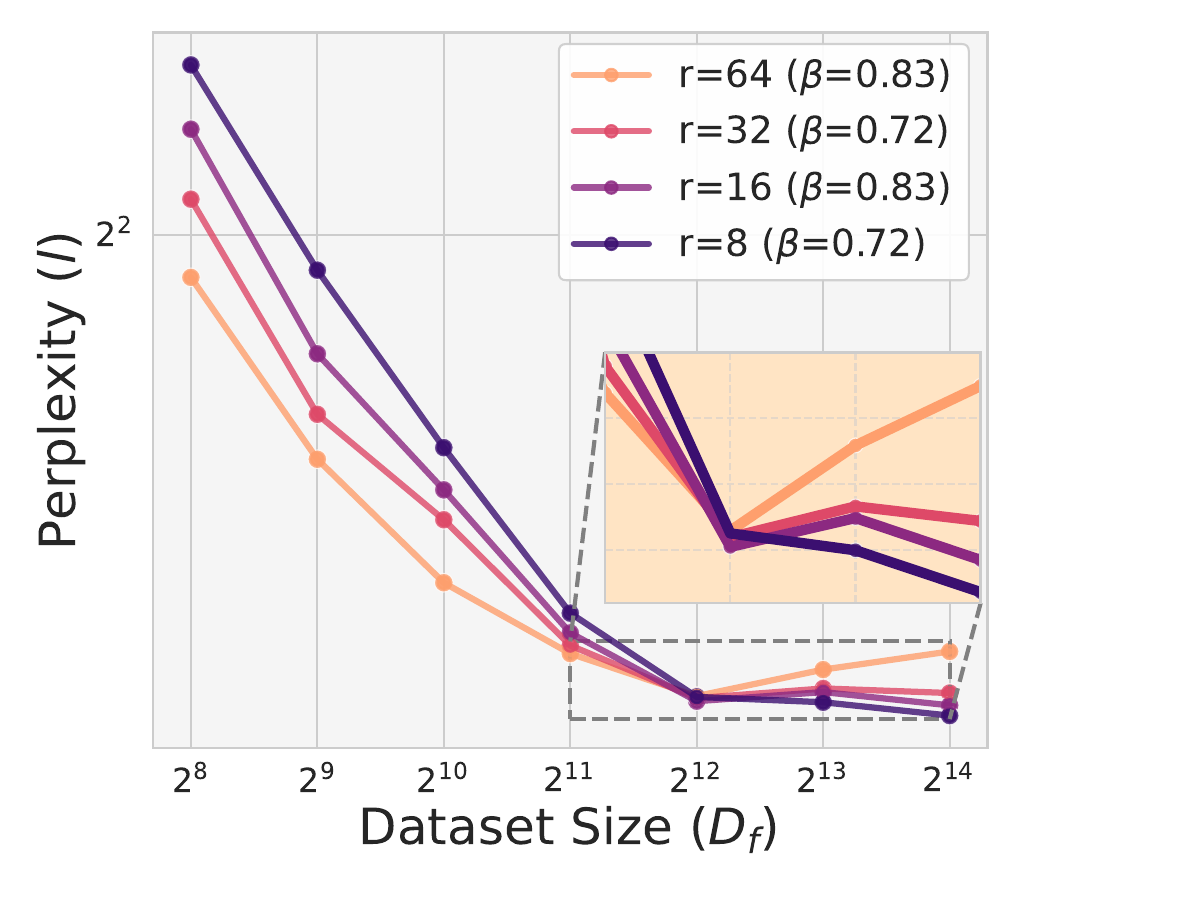}
          \caption{Scaling Law-P: Performance (perplexity) as a function of dataset size for different LLM ranks ($r_{\mathrm{ve}}=4$). The yellow line shows larger discrepancies between VE and LLM lead to reduced fine-tuning capability.}
          \label{fig:sc1} 
      \end{subfigure}
      \hfill
      \begin{subfigure}[t]{0.49\textwidth}
          \centering
          \includegraphics[width=\linewidth]{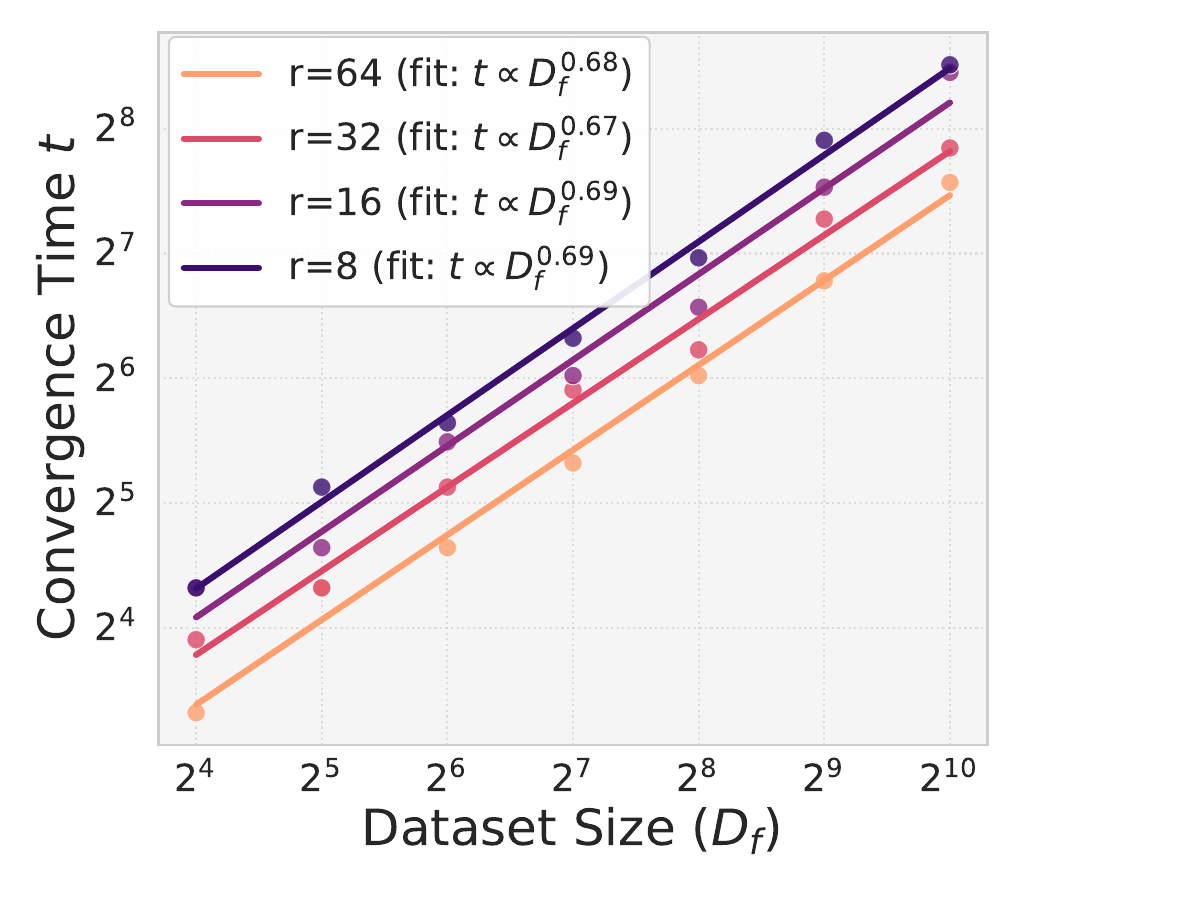}
          \caption{Scaling Law-C: Convergence time as a function of dataset size for different LLM ranks ($r_{\mathrm{ve}}=16$). The near-parallelism of the fitted lines supports the separability of the rank and dataset size terms in Equation~\ref{eq:scaling_law_2_general}.}
          \label{fig:sc2_b} 
      \end{subfigure}
  
      \caption{The proposed dual scaling laws from LLaVA-OV-0.5B* (VE and LLM have a similar parameter count without task-specific knowledge). This pattern holds consistently across different fixed-rank settings (e.g., $r_{\mathrm{ve}}=8$ and $r_{\mathrm{ve}}=32$).}
      \label{fig:sc2}
  \end{figure*}
  
  \subsubsection{Scaling Law-P for Performance} 
  \looseness=-1
  \paragraph{Formulation.} Based on our empirical findings and previous scaling laws on LLM fine-tuning~\cite{zhang2024sc_llmft}, we formulate Scaling Law-P to model the fine-tuning loss (or perplexity, $\hat{L}$) for a VLM. This law extends existing frameworks by incorporating separate LoRA ranks for the VE ($r_{\text{ve}}$) and the LLM ($r_{\text{llm}}$), alongside the dataset size ($D_f$):
  
  \begin{equation}
  \hat{L}(r_{\text{ve}}, r_{\text{llm}}, D_f) = A \cdot \frac{1}{(r_{\text{ve}})^{\alpha_m} \cdot (r_{\text{llm}})^{\alpha_l} \cdot D_f^{\beta}} + E
  \label{eq:scaling_law_p_main}
  \end{equation}
  
  \looseness=-1
  Here, $\alpha_m$ and $\alpha_l$ are scaling exponents for the VE and LLM ranks, reflecting their impact on loss reduction; $\beta$ is the exponent for dataset size. $A$ and $E$ are fitted constants representing a scaling coefficient and an irreducible error, respectively. This law serves as the objective function in MARS.
  
  
  \paragraph{Key Observation.}  
  Scaling Law-P is grounded in the foundational scaling laws~\cite{zhang2024sc_llmft}, which were
  originally established in the context of unimodal (LLM) fine-tuning. While this law holds under its original
  conditions, our work identifies a critical counterexample: \textbf{\textit{the law breaks down when there is a non-negligible imbalance
  in training dynamics across heterogeneous components.}}
  
  \begin{keyfindingbox_sl}
  \textbf{1) Performance is sensitive to the interplay between VE and LLM ranks.} Unlike unimodal studies, where LoRA rank often has a modest effect, VLM fine-tuning performance strongly depends on the combination of ranks chosen for the VE and the LLM. An imbalance between these ranks can undermine training benefits, particularly on large datasets (Figure~\ref{fig:sc1}), highlighting the need to consider both ranks jointly for effective fine-tuning.
  \end{keyfindingbox_sl}
  
  \vspace{2pt}
  \begin{keyfindingbox_sl}
  \textbf{2) Optimal LoRA rank pair involves a trade-off with dataset size.} Performance is non-monotonic: small datasets benefit from higher VE ranks for richer feature extraction (see Appendix~\ref{app:small-data-regime}), while large datasets favor moderate ranks to avoid overfitting. Thus, the optimal rank pair balances expressive feature learning with overfitting risk, depending on dataset size.
  \end{keyfindingbox_sl}
  
  These findings highlight the need for an adaptive, modality-aware rank search. Additional results, including counterexamples and justifications (e.g., why a multiplicative rather than additive formulation), are provided in Appendix~\ref{app:detailed_sc_p}.

  
  \subsubsection{Scaling Law-C for Convergence Time}
  \label{ssec:scaling_law_2_convergence}
  
  \paragraph{Formulation.}
  \label{sssec:scaling_law_2_formulation}
  Modeling the relationship between LoRA rank, dataset size, and convergence time (i.e., the number of iterations required for convergence) is a relatively underexplored area. To address this, we propose Scaling Law-C to model the convergence time for each modality-specific module within a VLM. 
  For a given module $i$, where $i$ represents either the VE or the LLM, we define the convergence time $t_i$ using the following general form:
  
  \begin{equation}
  t_{i}(r_{i}, D_f) = k_i \cdot (r_{i})^{\gamma_i} \cdot D_f^{\delta_i} + E_{i} \label{eq:scaling_law_2_general}
  \end{equation}
  
  \looseness=-1
  Here, $t_i$ is the predicted number of training steps to convergence for module $i$ (e.g., $i \in \{\text{ve}, \text{llm}\}$), with LoRA rank $r_i$ and dataset size $D_f$. The exponent $\gamma_i < 0$ captures the effect of rank on convergence time, while $\delta_i$ reflects the impact of dataset size. Constants $k_i$ and $E_i$ are fitted from data. This model predicts how convergence scales with rank and data, enabling MARS to balance $t_{\text{ve}}$ and $t_{\text{llm}}$ during pruning.

  \paragraph{Key Observation.}
  \label{sssec:scaling_law_2_observations}
  
  Our empirical results strongly support Scaling Law-C, showing consistent convergence time trends across varying dataset sizes and LoRA ranks. 
  
  \begin{keyfindingbox_sl}
  \textbf{1) Increasing dataset size \textit{increases} convergence time.} Larger fine-tuning datasets ($D_f$) require more iterations for the model to converge. This is intuitive, as the model needs additional steps to process and learn from the extra data. The near-linear trend, shown in Figure~\ref{fig:sc2_b}, aligns with the $D_f^{\delta_i}$ term in Equation~\ref{eq:scaling_law_2_general}.
  
  \end{keyfindingbox_sl}
    \vspace{2pt}
  
  \begin{keyfindingbox_sl}
  \textbf{2) Increasing rank size \textit{decreases} convergence time.} Modules with higher LoRA ranks ($r_i$) converge in fewer iterations, as larger ranks provide greater capacity and flexibility for efficient adaptation. Consequently, these modules reach convergence faster than those with lower, constrained ranks. This inverse relationship aligns with the $r_i^{\gamma_i}$ term in Equation~\ref{eq:scaling_law_2_general}, with negative $\gamma_i$ capturing the acceleration.
  \end{keyfindingbox_sl}
  
  These two distinct scaling behaviors form the empirical basis for our convergence time predictions, which are integral to MARS's ability to identify convergence-balanced rank pairs. Additional \textit{theoretical} and \textit{empirical} evidence is provided in Appendix~\ref{app:detailed_sc_c}.

  \subsection{The Guided Search Space Optimization}
  \label{sec:search}
  
  \looseness=-1
  Once the dual scaling laws are calibrated, MARS performs an automated search for the optimal LoRA rank pair $(r^*_{\text{ve}},~r^*_{\text{llm}})$ for a target dataset size.
  The search first prunes the vast space of possible rank combinations using our fitted Scaling Law-C.  This strategy is computationally grounded in the \textbf{\textit{monotonic smoothness}} and \textbf{\textit{local continuity}} of the scaling landscape observed in our evidence (see Figure~\ref{fig:sc2_b} \& Appendix~\ref{app:detailed_sc_c}). Because the training dynamics are smooth and order-preserving without abrupt spikes or discontinuities across the search space, MARS models Scaling Law-C as a continuous power-law function, enabling efficient pruning via interpolation between discrete anchor points.
  
  \looseness=-1
  The guiding principle is to identify rank pairs where the VE and LLM modules achieve aligned convergence times ($t_{\text{ve}}~\approx~t_{\text{llm}}$). By leveraging the overall smoothness of these scaling curves, we can solve this balance condition to express the ideal VE rank as a direct function of the LLM rank:
  
  \begin{equation}
  r_{\text{ve}} \approx \left( \frac{k_l \cdot (r_{\text{llm}})^{\gamma_l} \cdot D_{f}^{\delta_l} + E_{\text{llm}} - E_{\text{ve}}}{k_v \cdot D_{f}^{\delta_v}} \right)^{\frac{1}{\gamma_v}}
  \label{eq:r_ve_ideal}
  \end{equation}

  To generate a set of \textit{convergence-aligned} candidates, MARS iterates through a list of possible LLM ranks, uses Equation~\ref{eq:r_ve_ideal} to calculate the corresponding ideal VE rank for each, and rounds to the closest integer based on the local continuity and smoothness of the scaling landscape. This procedure efficiently creates a pruned set of candidate pairs, $CandidatePairs$, in Algorithm~\ref{alg:mars}. See Appendix~\ref{app:derivation} for the full derivation.

  The final step is to select the pair with the highest performance potential. We use the fitted Scaling Law-P to predict the fine-tuning perplexity, $\hat{L}$, for each candidate pair, with $\Theta_{\text{acc}}$ denoting the fitted coefficients. The optimal rank pair is the one that minimizes this predicted perplexity:
  
  \begin{equation}
  (r^*_{\text{ve}}, r^*_{\text{llm}}) = \underset{(r_{\text{ve}}, r_{\text{llm}}) \in \texttt{CandidatePairs}}{\text{argmin}} \hat{L}(r_{\text{ve}}, r_{\text{llm}}, D_f; \Theta_{\text{acc}})
  \label{eq:argmin_loss}
  \end{equation}
  
  The value of our dual scaling law framework lies in its synergistic two-phase approach. A search guided only by a performance predictor can yield \textit{practically unstable} configurations due to the gradient interference caused by imbalanced convergence, while optimizing only for balance does not guarantee the highest accuracy. MARS resolves this trade-off by first using Scaling Law-C to prune the search space to a stable subset of candidates, and then employing Scaling Law-P to select the highest-performing solution from that set.
  
  \subsection{Calibration of Dual Scaling Laws}
  \label{sec:calibration}
  \looseness=-1
  When fine-tuning begins, the MARS procedure starts with a calibration phase that empirically estimates the coefficients of our dual scaling laws. To minimize computational cost, we employ an efficient data collection strategy: for each representative LoRA rank pair (e.g., $r_{\text{ve}}, r_{\text{llm}} \in \{8, 16, 32, 64\}$), we initiate a single fine-tuning run and record performance and convergence metrics at multiple intermediate checkpoints corresponding to smaller effective dataset sizes (e.g., at $2^{10}$ and $2^{11}$ fine-tuning steps). This approach yields a rich dataset, $\mathcal{D}$, from a minimal number of training runs. Using this dataset, we determine the coefficients for both scaling laws by framing it as a numerical optimization problem. 
  Following established practices~\cite{hoffmann2022hubber, zhang2024sc_llmft}, we employ the L-BFGS-B algorithm to find the parameter set that minimizes the Huber loss between our model's predictions and the observed data. 
  Crucially, this calibration stage utilizes the same fine-tuning settings (e.g., optimizer, batch size) as the full fine-tuning, ensuring the identified rank pair remains effective. The result is a reliable predictive model of performance and convergence dynamics, essential for the subsequent guided search.

  \section{Evaluation}
  \subsection{Evaluation Setup}
  \label{sec:eval_setup}
  \paragraph{Evaluation Focus.}
  To ensure a comprehensive assessment, we consider the overall phases of VLM fine-tuning in alignment with current community trends and established practices, which can be categorized into two types:
  
  \begin{itemize}
      \item \textbf{Generalist Fine-tuning.}
      Fine-tuning the model on a diverse set of general multimodal instructions to build a strong foundational generalist. This stage is often referred to as \emph{pre-training} after the pre-trained VE and LLM are first connected, as these components have not previously been exposed to multimodal tasks.
      \item \textbf{Specialist Fine-tuning.}
      Fine-tuning a pre-trained generalist model on domain-specific datasets to produce a specialized expert. This stage is generally referred to as \emph{post-training}. Given that the model possesses strong general knowledge, effective fine-tuning must balance acquiring task-specific knowledge while preserving pre-trained representations.
  \end{itemize}

  \begin{table}[t]
  \centering
  \footnotesize
  \begin{minipage}[t]{0.49\linewidth}
  \centering
  \caption{Comparison with fixed-rank tuning across different learning rates. Best in bold, second-best underlined. OV: LLaVA-OV; Qwen: Qwen2.5-VL.}
  \label{tab:comparison_lr_combined}
  \resizebox{\linewidth}{!}{%
  \renewcommand{\arraystretch}{1.3}
  \begin{tabular}{l  l ccc c}
  \toprule
  \textbf{Model} & \textbf{Bench.} & \multicolumn{3}{c}{\textbf{LoRA} ($\textbf{lr}_{\textbf{ve}}$, $\textbf{lr}_{\textbf{llm}}$)}  & \textbf{MARS} \\
  \cmidrule(lr){3-5}
  &  & ($\bigstar$, 1e-5) & ($\bigstar$, 1e-6) & ($\bigstar$, 1e-7) & \\
  \midrule
  \multirow{2}{*}{\shortstack{OV-0.5B\\ \cite{li2024llava-ov}}}  & LLaVA ($\downarrow$) & \underline{2.7336}	& 2.771	& 2.8472	& \textbf{2.7188}\\
  & SciQA ($\uparrow$) & \underline{71.06}  & 61.88 & 59.28 & \textbf{72.85}\\
  \multirow{2}{*}{\shortstack{OV-7B\\ \cite{li2024llava-ov}}}  & LLaVA ($\downarrow$) & 2.2317	& \underline{2.295}	& 2.4346	& \textbf{2.1875} \\
  & SciQA ($\uparrow$) & \underline{72.26} & 69.86  & 67.27 & \textbf{74.25} \\
  \multirow{2}{*}{\shortstack{Qwen-3B\\ \cite{bai2025qwen25}}}  & LLaVA ($\downarrow$) & \underline{3.6156}	& 3.7415	& 4.1557	& \textbf{3.5925} \\
  & SciQA ($\uparrow$) & \underline{78.04} & 76.45 & 76.25 & \textbf{79.24} \\
  \multirow{2}{*}{\shortstack{Qwen-7B\\ \cite{bai2025qwen25}}}  & LLaVA ($\downarrow$) & \underline{3.5032}	& 3.5908	& 3.8716	& \textbf{3.3879}\\
  & SciQA ($\uparrow$) & \textbf{79.84} & 76.25 & 74.25 & \underline{79.64} \\
  \bottomrule
  \end{tabular}}
  \end{minipage}
  \hfill
  \begin{minipage}[t]{0.49\linewidth}
  \centering
  \caption{Comparison with adaptive rank search baselines. Variance is low across three seeds (std < 0.3\%); See Appendix~\ref{sec:stability} for stability details; AdaL.: AdaLoRA; GeoL: GeoLoRA.}
  \label{tab:comparison_baselines_mars}
  \resizebox{\linewidth}{!}{%
  \renewcommand{\arraystretch}{1.2}
  \begin{tabular}{lccccccc}
  \toprule
  \textbf{Model} & \multirow{2}{*}{\textbf{AdaL.}} & \multirow{2}{*}{\textbf{GeoL.}} &\textbf{Full-rank} & \multicolumn{3}{c}{\textbf{LoRA} ($\textbf{r}_{\textbf{ve}}$, $\textbf{r}_{\textbf{llm}}$)}  & \multirow{2}{*}{\textbf{MARS}} \\
  \cmidrule(lr){5-7}
  &  & &  \textbf{Tuning} & ($\bigstar$, 8) & ($\bigstar$, 16) & ($\bigstar$, 32) &  \\
  \midrule
  \multicolumn{8}{l}{\textit{LLaVA Bench (perplexity $\downarrow$)} }  \\
  \midrule
  OV-0.5B & 2.8973  & 2.8801  & \underline{2.7209}   & 2.7582    & 2.7336           & 2.7331        & \textbf{2.7188}       \\
  OV-7B &  2.5189   & 2.4888 & 2.2693     & 2.3181      & \underline{2.2317}     & 2.4420         & \textbf{2.1875}              \\
  Qwen-3B  & 3.6679 & 3.6676  & 3.6528     & \underline{3.5912}      & 3.6156   & 3.7811          & \textbf{3.5825}   \\
  Qwen-7B  & 3.6394  & 3.6378 & 3.5917         & \underline{3.5011}        & 3.5032         & 3.6098              & \textbf{3.3616}  \\
  \midrule
  \multicolumn{8}{l}{\textit{ScienceQA bench (accuracy (\%) $\uparrow$)}} \\
  \midrule
  OV-0.5B &  62.28 & 63.52 &69.66 & 70.46& \underline{71.06} & 69.86  & \textbf{72.85} \\
  OV-7B & 66.27 & 67.81 & 70.46 & 70.26 & 72.26  & \underline{73.65}  & \textbf{74.25} \\
  Qwen-3B  & 70.06 & 70.46 & 75.25 & \underline{78.24} & 78.04& 78.04  & \textbf{79.24} \\
  Qwen-7B  & 73.85 & 72.85 & 77.45 & \textbf{79.84} & 79.84 & 78.24 & \underline{79.64} \\
  \bottomrule
  \end{tabular}}
  \end{minipage}
  \end{table}
  
  \paragraph{Evaluation Benchmarks and Datasets.}
  \looseness=-1
  To evaluate the effectiveness of MARS, we utilize two distinct benchmarks designed to measure different types of fine-tuning performance.
  For generalist fine-tuning, we use LLaVA Bench\cite{liu2023llava}, which consists of 15k validation samples from the LLaVA-Instruct dataset and is composed of complex reasoning tasks. For this benchmark, we use validation perplexity as the primary metric. 
  Perplexity, computed directly from cross-entropy loss, serves as a proxy for the model’s training objective (minimizing loss) and provides a direct measure of its fine-tuning capability by indicating how well it can specialize on a downstream dataset.
  For specialist fine-tuning, we use the ScienceQA benchmark~\cite{lu2022scienceqa}. This benchmark contains multimodal multiple-choice questions across diverse scientific subjects. 
  For our experiments, we use a 16k training split and evaluate on 500 test samples. Following standard convention, we report accuracy. We further extend MARS to a broader set of benchmarks to evaluate its generality:
  MME~\cite{fu2023mme}, MMStar~\cite{chen2024mmstar}, POPE~\cite{pope}, TextCaps~\cite{sidorov2019textcaps}, AI2D~\cite{ai2d}, GQA~\cite{hudson2018gqa}. Benchmark details are provided in Appendix~\ref{appendix: generality_appendix_section}.
  
  \paragraph{Baselines.}
  To demonstrate the effectiveness of MARS as an adaptive rank search algorithm for convergence coordination, 
  we compare it against several key baselines. The first is a differential learning rate approach, 
  representing the standard heuristic of manually tuning separate learning rates for the VE and LLM. 
  Second, since we argue that using differential ranks between modalities is important, 
  we compare against a set of fixed differential rank pairs (i.e., uniform but different across modalities). 
  Third, we benchmark against AdaLoRA~\cite{zhang2023adalora} and GeoLoRA~\cite{schotthofer2024geolora},
  two representative adaptive rank allocation methods designed for unimodal models (see Appendix~\ref{app:adalora} for details). 
  Finally, we compare against two representative search methods: 
  Random Search and Bayesian Optimization (TPE~\cite{watanabe2023tpe}), where a probabilistic model guides the search.
  By comparing against these methods, we demonstrate that MARS effectively identifies the optimal rank combination.
  
  \paragraph{Implementation Details.}
  Our experiments are implemented using the Cornstarch MLLM training framework~\cite{jang2025cornstarch}. All models were fine-tuned using the Adam optimizer with a cosine learning rate scheduler, and all experiments were conducted on a single NVIDIA GH200 GPU. 
  Further details on hyperparameters, LoRA settings, and the MARS calibration setup are provided in the Appendix~\ref{app:implementation_details}.

  
  \subsection{Main Results}
  \label{ssec:initial_mars_result}
\begin{figure}[t]
      \centering
      \begin{minipage}[c]{0.56\linewidth}
          \centering
          \includegraphics[width=0.95\linewidth]{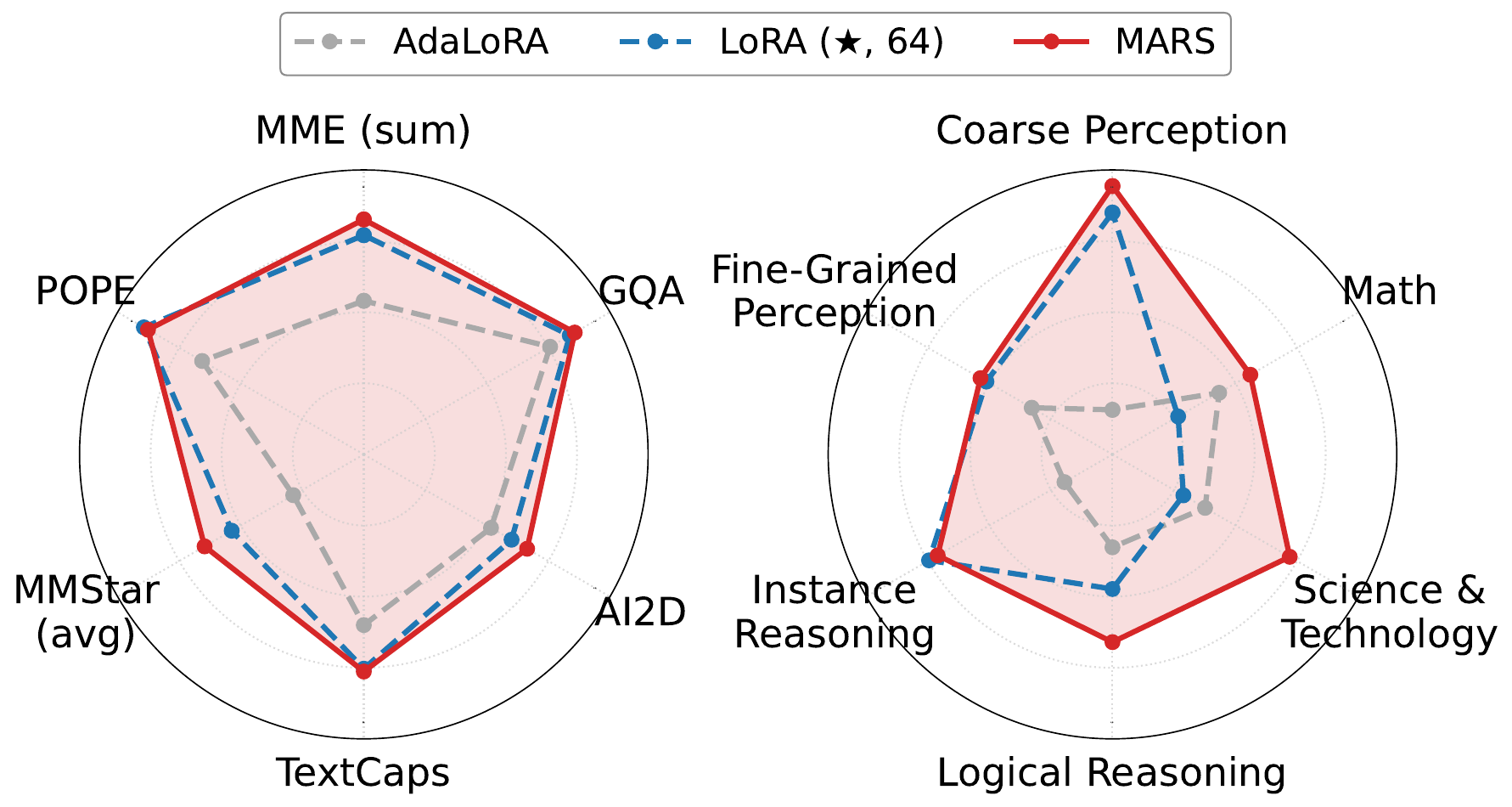}
          \captionof{figure}{Generalist Capabilities: \textbf{(L)}~Comparison across diverse multimodal benchmarks. \textbf{(R)} Fine-grained breakdown on MMStar. Details in Appendix~\ref{appendix: generality_appendix}.}
          \label{fig:generality}
      \end{minipage}
      \hfill
      \begin{minipage}[c]{0.40\linewidth}
          \centering
          \vspace{-28pt}
          \captionof{table}{Correlation between Convergence Gap ($|T_{ve} - T_{llm}|$) and Perplexity.}
          \label{tab:convergence_stat}
          \resizebox{0.7\linewidth}{!}{%
          \begin{tabular}{@{}lcc@{}}
              \toprule
              \textbf{Data Size} & \textbf{Pearson's $r$} & \textbf{P-value} \\
              \midrule
              0.5K  & 0.90 & 0.0020 \\
              1K & 0.95 & 0.0003 \\
              2K & 0.87 & 0.0052 \\
              4K & 0.98 & 0.0000 \\
              \bottomrule
          \end{tabular}%
          }
          \includegraphics[width=0.75\linewidth]{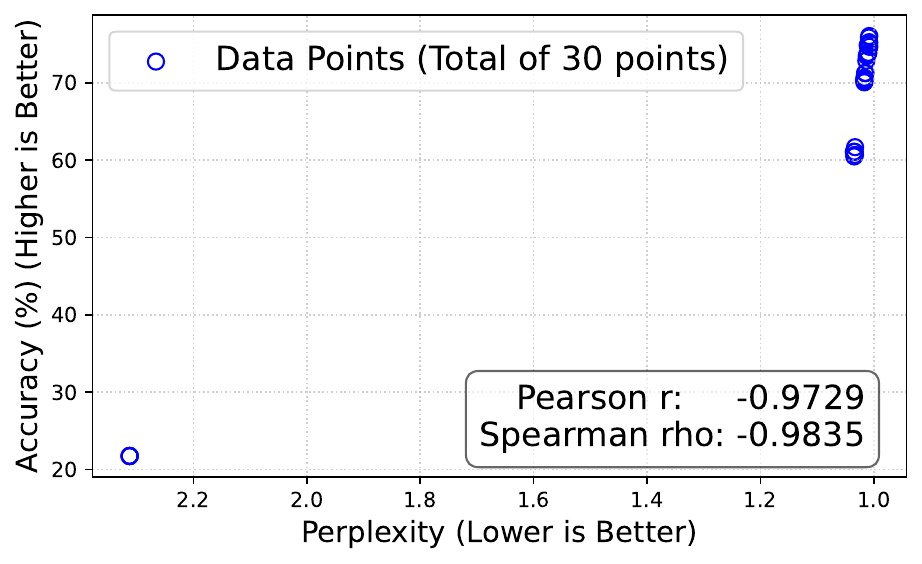}
          \captionof{figure}{Correlation between Perplexity and Task Accuracy.}
          \label{fig:scienceqa_correlation}
      \end{minipage}
  \end{figure}

  We evaluate MARS across multiple VLM architectures and model scales, comparing it against several strong baselines across various benchmarks. As shown in Tables~\ref{tab:comparison_lr_combined} and~\ref{tab:comparison_baselines_mars} and Figure~\ref{fig:generality}, the fine-tuned models with MARS show effectiveness and generality compared to our baselines.
  
  \paragraph{Comparison with Differential Learning Rates.}
  Table~\ref{tab:comparison_lr_combined} compares MARS against the common heuristic of using fixed LoRA ranks while tuning differential learning rates. On the LLaVA Bench, MARS consistently achieves a lower (better) perplexity across all tested models. For instance, on the LLaVA-OV-7B model, MARS achieves a perplexity of 2.1875, outperforming the best-performing differential LR setting (2.295). This trend holds on the more specialized ScienceQA benchmark, where MARS improves the accuracy of the LLaVA-OV-7B model from 72.26\% to 74.25\%. These results demonstrate that our rank-centric search is a more effective strategy for harmonizing module dynamics than just tuning learning rates.
  See Appendix~\ref{app:from-scratch} for results on the from-scratch models.

  \paragraph{Comparison with Other Baselines.}
  In Table~\ref{tab:comparison_baselines_mars}, we benchmark MARS against a wider range of fine-tuning strategies. 
  MARS consistently surpasses Full-rank Tuning, highlighting the regularization benefits of our adaptive PEFT approach. 
  More importantly, it outperforms various fixed differential LoRA rank pairs, proving the value of our guided search over 
  simply selecting arbitrary rank combinations. 
  The most critical comparison is with AdaLoRA and GeoLoRA, two representative adaptive-rank methods designed for unimodal models. 
  We evaluate them under the equal-budget setting as MARS-selected ranks.
  Due to their algorithmic characteristics 
  (e.g., AdaLoRA starts with larger ranks and progressively prunes, while GeoLoRA dynamically expands and shrinks ranks), 
  smaller effective ranks are required to match the same budget. 
  Moreover, their layer-wise importance estimation is not optimal in VLM settings, leading both methods to significantly underperform.

\looseness=-1
\paragraph{Comparison with Equal-Budget Search Baselines.}
We compare MARS against equal-budget representative search methods under the equal trainable-parameter budget. 
As shown in Table~\ref{tab:equal_budget}, MARS outperforms both Random Search (best of 5 trials) and 
Bayesian Optimization (TPE with 6 trials) with lower search overhead. 
For AdaLoRA, we find that the per-module variant outperforms unified rank assignment.

\begin{table}[t]
  \centering
  \footnotesize
  \begin{minipage}[t]{0.57\linewidth}
    \centering
    \caption{Equal-budget comparison on Qwen2.5-VL-3B.  AdaLoRA (separate): separate per module; AdaLoRA (unified): treats VLM as unimodal.}
    \label{tab:equal_budget}
    \resizebox{\linewidth}{!}{%
    \begin{tabular}{lccccc}
    \toprule
    & & \multicolumn{2}{c}{\textbf{ScienceQA (Acc $\uparrow$)}} & \multicolumn{2}{c}{\textbf{LLaVABench (PPL $\downarrow$)}} \\
    \cmidrule(lr){3-4} \cmidrule(lr){5-6}
    \textbf{Method} & \textbf{Total Cost} & \textbf{Params (M)} & \textbf{Result} & \textbf{Params (M)} & \textbf{Result} \\
    \midrule
    Random Search & 5 full runs & 26--27 & 77.64\% & 22--24 & 3.5928 \\
    Bayesian (TPE) & 6 full runs & $\leq$27.19 & 78.64\% & $\leq$23.62 & 3.6882 \\
    AdaLoRA (separate) & 1 full run & 29.97 & 71.26\% & 22.79 & 3.6493 \\
    AdaLoRA (unified) & 1 full run & 25.86 & 70.06\% & 25.86 & 3.6679 \\
    MARS & Calib+1 full run & 27.19 & 79.24\% & 23.62 & 3.5825 \\
    \bottomrule
    \end{tabular}}
  \end{minipage}%
  \hfill
  \begin{minipage}[t]{0.4\linewidth}
    \centering
    \caption{Comparison against the exhaustive grid search optimum. ($\textbf{r}_{\textbf{ve}}$, $\textbf{r}_{\textbf{llm}}$).}
    \label{tab:mars_vs_grid}
    \resizebox{\linewidth}{!}{%
    \begin{tabular}{llcc}
    \toprule
    \textbf{Benchmark} & \textbf{Model} & \textbf{Exhaustive Top-1} & \textbf{MARS} \\
    \midrule
    LLaVAB. & LLaVA-OV-7B & (10, 16) & (10, 16) \\
    LLaVAB. & Qwen2.5VL-3B & (10, 8) & (10, 8) \\
    ScienceQA & LLaVA-OV-7B & (24, 32) & (22, 32) \\
    ScienceQA & Qwen2.5VL-3B & (15, 8) & (15, 8) \\
    \bottomrule
    \end{tabular}}
  \end{minipage}
  \end{table}

\paragraph{Comparison with Exhaustive Grid Search.}
To verify that the convergence-alignment pruning does not exclude high-performing candidates, we compare MARS against a 64-point exhaustive grid search ($r_{\text{llm}} \in \{4,8,16,32\} \times r_{\text{ve}} \in \{2,4,\ldots,32\}$). Table~\ref{tab:mars_vs_grid} shows that MARS’s scaling-law-based alignment effectively identifies or closely approximates optimal rank pairs.

  \paragraph{Generality with Broader Benchmark Coverage.}
  
  To demonstrate the generality of MARS, we evaluate the model on a diverse set of benchmarks, including both generic and domain-specific tasks. Given the limited data diversity of the initial LLaVA dataset~\cite{liu2023llava}, we use the updated version, LLaVA-1.5-665K~\cite{liu2024llava15}, to train LLaVA-OV-7B$^\ast$. This experiment is generalist fine-tuning with a from-scratch model, isolating pretrained knowledge to assess MARS's generality. As shown in Figure~\ref{fig:generality}, MARS mostly outperforms the strongest LoRA baselines (rank combinations $r_{\text{ve}} \in \{32,48,64\}$, $r_{\text{llm}}=64$) and adaptive methods such as AdaLoRA (target rank= 64) across the benchmarks. For this experiment, we set $R_{\text{options}}$ (i.e., candidate ranks for the LLM) to $\{32, 48, 64\}$ to better absorb the knowledge from the large, diverse dataset.
  
  \begin{figure}[tp]
      \centering
      \begin{minipage}[c]{0.70\linewidth}
          \centering
          \includegraphics[width=0.9\linewidth]{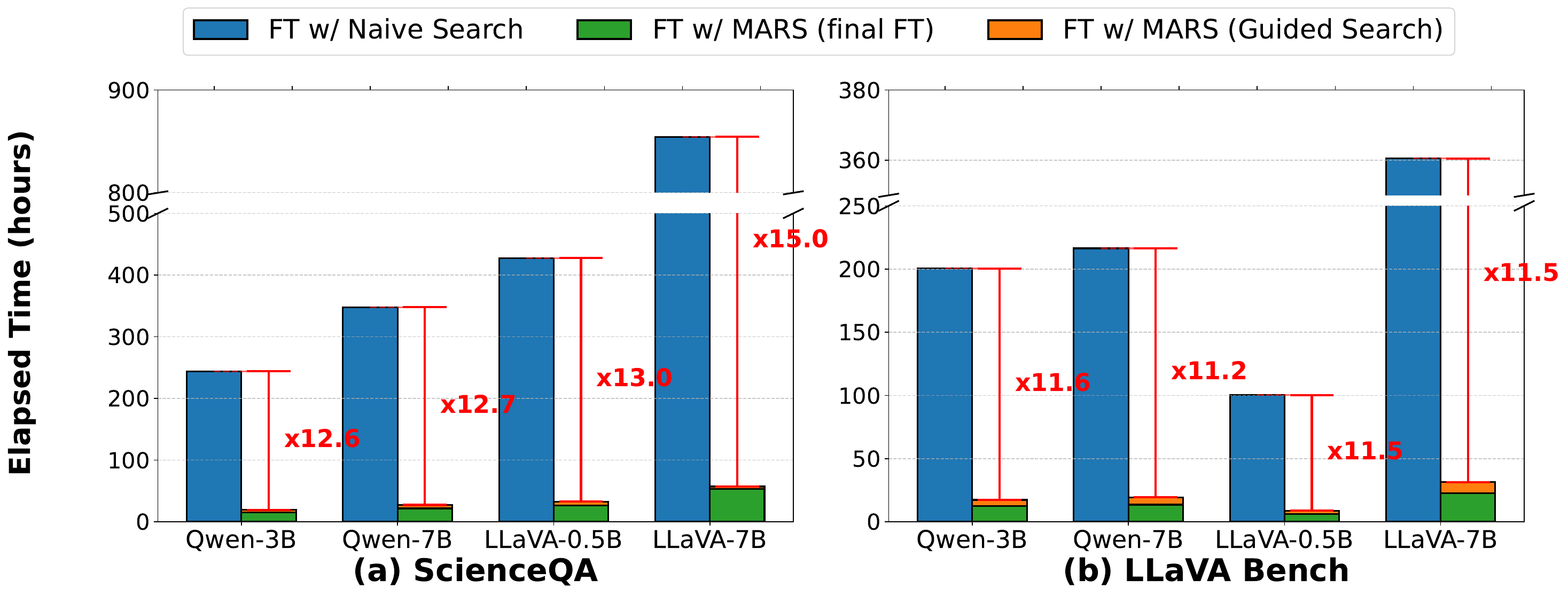}
          \captionof{figure}{Time comparison with Naive Search. Naive Search explores the grid $\{4, 8, 16, 32\} \times \{4, 8, 16, 32\}$ to find the best rank pair.}
          \label{fig:gputime}
      \end{minipage}
      \hfill
      \begin{minipage}[c]{0.28\linewidth}
          \centering
          \vspace{6pt}
          \includegraphics[width=\linewidth]{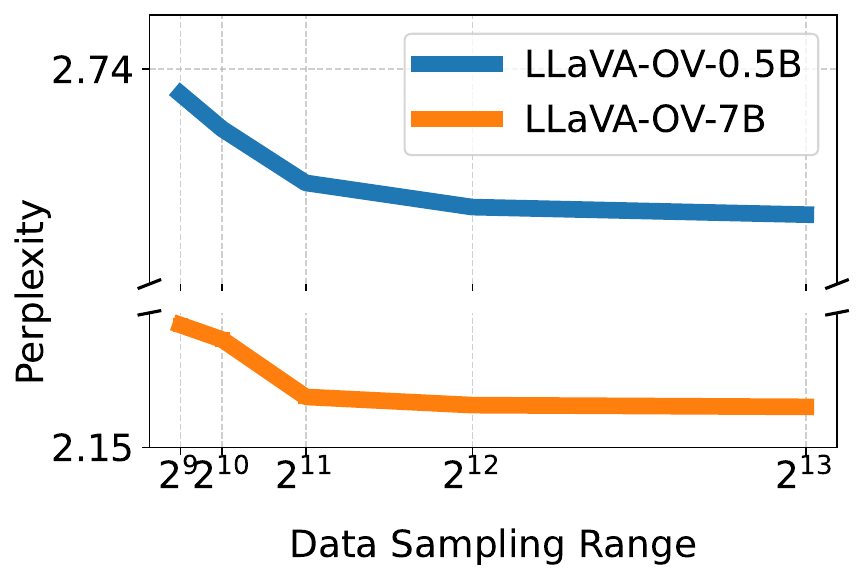}
          \vspace{3pt}
          \captionof{figure}{Impact of Data Sampling Range.}
          \label{fig:datasampling}
      \end{minipage}
  \end{figure}
  \subsection{Computational Efficiency}
  Beyond its advantages in fine-tuning effectiveness, MARS also achieves a substantial improvement in computational efficiency. As shown in Figure~\ref{fig:gputime}, a naive exhaustive search for the optimal rank pair requires extensive compute, often exceeding 100 GPU hours. In contrast, MARS consists of a lightweight calibration phase followed by a single fine-tuning run. This reduces the total elapsed time by more than 11.5x on average across different models and tasks, with most of the time spent on the actual fine-tuning (which is unavoidable). This substantial reduction in search cost demonstrates that MARS is not only more effective, but also more practical and scalable for VLM fine-tuning.

  \subsection{Impact of Convergence Alignment on Performance}
  
  To validate the hypothesis that aligning the convergence times of the Vision Encoder (VE) and the Large Language Model (LLM) leads to better final performance, we performed a two-part statistical analysis linking convergence synchronization, perplexity, and final accuracy.
  
  \textbf{Perplexity as a Performance Proxy.}
  First, we establish the reliability of perplexity as a robust, task-agnostic proxy for fine-tuning quality. 
  While prior studies~\cite{li2024llava-ov, shukor2025sc_mm} already suggest a link between lower perplexity and improved downstream performance,
  we empirically validate this relationship in our fine-tuning setup using the ScienceQA dataset. 
  Figure~\ref{fig:scienceqa_correlation} shows a strong negative correlation between perplexity and task accuracy (Pearson’s $r = -0.97$, $p < .001$), 
  confirming that optimizing the perplexity proxy effectively maximizes task accuracy.
  This relationship generalizes across six diverse benchmarks (Pearson $r$ from $-0.72$ to $-0.97$; 
  Appendix~\ref{app:ppl_correlation}).

  \textbf{Linking Convergence Gap to Performance.}
  Building on this proxy, we directly address the relationship between the \textit{Convergence Gap} ($|T_{ve} - T_{llm}|$) and model perplexity. We conducted a statistical correlation analysis across multiple fine-tuning regimes with dataset sizes ranging from $D = 2^9$ to $2^{13}$. As in Table~\ref{tab:convergence_stat}, we observed statistically significant positive correlations ($r > 0.86$, $p < 0.01$) across all scales. These results confirm that minimizing the convergence gap is a statistically dominant predictor of reduced perplexity and, by extension, superior model performance.

  \begin{wraptable}{r}{0.43\linewidth}
    \vspace{-14pt}
    \centering
    \footnotesize
    \caption{Prediction reliability of Scaling Law-P on pre-trained LLaVA-OV. $R^2$: fit quality; MAPE: mean abs. percentage error.}
    \label{tab:calibration_r2}
    \begin{tabular}{lcc}
    \toprule
    \textbf{Source of the coefficients} & \textbf{$R^2$} & \textbf{MAPE} \\
    \midrule
    From-scratch LLaVA-OV & $-5.309$ & 57.23\% \\
    Per-model calibration & $0.994$ & 0.96\% \\
    \bottomrule
    \end{tabular}
  \end{wraptable}

  \subsection{Functional-Form Transferability.}
   Table~\ref{tab:calibration_r2} tests whether per-model calibration is necessary and sufficient. Reusing coefficients from a different model leads to a severe mismatch, whereas calibrating on the target model achieves a strong fit. These results suggest that the functional form of the scaling law is transferable across regimes, but the coefficients must be recalibrated for each model, as assumed in MARS (see Appendix~\ref{sec:calibration_analysis} for further analysis).

  \subsection{Ablation: Impact of Data Sampling Range.}
  We conducted an ablation study to determine the optimal range of data required for the lightweight calibration phase of MARS. We ran the MARS search procedure multiple times, each time using a different maximum data sampling range for calibration, and then evaluated the final test perplexity achieved by the resulting optimal rank pair. As shown in Figure~\ref{fig:datasampling}, we observe that increasing the data sampling range up to $2^{11}$ leads to a clear improvement in the final perplexity, as it allows for a more accurate fitting of the scaling laws. However, performance gains diminish significantly beyond that point, validating our choice of using a data sampling range up to $2^{11}$ for MARS calibration, as it strikes an effective balance between accuracy and computational efficiency.

  \section{Discussion}
  \label{sec:discussion}
  
  \paragraph{Scalability with Modality Expansion.}
  The current scope of MARS focuses on VLMs, while tri-modal settings (vision–audio–language~\cite{2026qwen-omni}) are left for future work.
  However, we believe MARS can help address the combinatorial explosion of multimodal hyperparameter tuning. In contrast to the exponential growth ($O(C^N)$) of naive grid search, MARS transforms the search into $(N-1)$ independent 1D equations by anchoring the LLM rank as a reference. Furthermore, to minimize the overhead of the calibration phase, we employ a Simultaneous Multi-Rank Adapter Tuning strategy (see Appendix~\ref{appendix:scalability}). By leveraging a frozen backbone with multiple parallel adapters, we reduce the calibration cost to near-constant time $O(1)$, ensuring that MARS remains computationally efficient.
  

  \paragraph{Heterogeneous Gains.}
  MARS’s gains vary with the degree of convergence imbalance. When the optimal rank pair coincides with a standard configuration (e.g., $r_{\text{ve}} = r_{\text{llm}} = 8$ for Qwen2.5-VL-7B on ScienceQA in Table~\ref{tab:comparison_baselines_mars}), improvements over exhaustive search are marginal. In such cases, MARS’s main benefit is efficiency, identifying optimal configurations without $>$11$\times$ the cost of grid search.



  \section{Conclusion}
  \label{sec:conclusion}
  \looseness=-1
  We introduce MARS to address imbalanced convergence in VLM fine-tuning. By leveraging dual scaling laws to predict convergence and prune the LoRA rank search space, MARS identifies optimal rank pairs with balanced dynamics and outperforms common heuristic-based strategies, highlighting the importance of convergence coordination for effective multimodal adaptation.
  As future work, we aim to systematically characterize fine-grained relationships between domain sensitivities across tasks.

  \bibliographystyle{unsrt}
  \bibliography{references}

  \newpage
  \appendix
  \onecolumn
  \section*{Appendix}
  
  \section{Additional Experimental Details for Section~\ref{sec:dual_scaling_laws}}
  \label{app:exp_details}
  
  \subsection{Model Setup}
  \label{app:model_setup}
  
  Our target model architecture is LLaVA-OneVision (LLaVA-OV;~\cite{li2024llava-ov}). For these scaling law experiments, all models were initialized ``from scratch,'' a methodology we denote with an asterisk (*). This process involves downloading publicly available, pre-trained vision encoder (VE) and Large Language Model (LLM) checkpoints and connecting them with an MLP projector, without any subsequent instruction tuning. Specifically, we used SigLIP~\cite{zhai2023siglip} for the VE and Qwen2~\cite{team2024qwen2llm} for the LLM and attached a 2-layer MLP projector to replicate the original structure.
  
  This approach is intentionally designed to ensure that the model's initial state is \textit{domain-knowledge-free} with respect to multimodal instruction data. By doing so, we can isolate and measure the effects of pure fine-tuning capability as a function of data and model scale. We investigated two architectural variants to analyze the impact of the parameter gap between the VE and LLM components.
  
  \begin{itemize}
      \item \textbf{LLaVA-OV-0.5B*:} Features a \textit{minimal} parameter gap (VE $\sim$400M, LLM $\sim$500M).
      \item \textbf{LLaVA-OV-7B*:} Features a \textit{significant} parameter gap (VE $\sim$400M, LLM $\sim$7B).
  \end{itemize}
      

  \subsection{Dataset Setup}
  \label{app:dataset_setup}
  Our primary dataset is the LLaVA dataset~\cite{liu2023llava,li2024llava-ov}, which, to our knowledge, is one of the largest of its kind available for research. Each sample is centered on images of nature or everyday life. The dataset comprises a mixture of open-ended task types, including basic image captioning, detailed image captioning, and multi-turn conversational question answering.
  
  To study scaling trends, we constructed training subsets from this dataset by varying their size according to a power-of-two progression (i.e., $2^n$ samples, for $n$ from 3 to 15), while maintaining a consistent distribution of the above task types across all subsets. For validation, we used a fixed subset of 500 samples (non-overlapping with training data) to enable frequent, low-cost perplexity checks during intermediate training cycles.

  \section{Additional Related Work}
  \label{app:adalora}
  \paragraph{Adaptive LoRA for Unimodal Models.}
  To move beyond static configurations, adaptive LoRA methods like AdaLoRA~\cite{zhang2023adalora}, SalientLoRA~\cite{ke2024unveiling}, and GeoLoRA~\cite{schotthofer2024geolora} were developed. However, these methods are designed for unimodal models and face fundamental limitations in multimodal setups. In the unimodal context, it is often observed that allocating a higher rank to more salient layers correlates with improved transferability and performance~\cite{biderman2024loralearnless}, paralleling model compression techniques that allocate more resources to higher-importance components~\cite{dong2019hawq, lee2021novel}. In contrast, we found that for VLMs, simply assigning higher ranks does not invariably lead to better performance. This is due to the complex and crucial inter-modality dependencies, where the optimal configuration for one modality is influenced by the state of the other.
  
  
  \section{MARS Implementation Details}
  \label{app:implementation_details}
  
  \subsection{General Training Hyperparameters}
  
  Unless otherwise specified, all fine-tuning experiments were conducted with the following hyperparameters. We used the Adam optimizer with its default parameter settings and a batch size of 8. The learning rate was initialized to $1 \times 10^{-5}$ and followed a cosine decay schedule with a warmup phase corresponding to 10\% of the total training steps. These settings were applied consistently across all experiments, with the exception of the peak learning rate in the baseline tests for differential learning rates. Following a prior paper~\cite{liu2023llava}, we trained for 3 epochs on the LLaVA task and 12 epochs on the ScienceQA task, reporting the best performance.
  
  \subsection{LoRA Configuration}
  For all experiments utilizing LoRA, we applied LoRA to all linear layers across the entire model, excluding the output heads. The LoRA scaling hyperparameter, $\alpha$, was consistently set to twice the rank size (i.e., $\alpha = 2 \times r$)~\cite{liu2024dora}. The initial calibration phase for MARS was performed using a representative set of LoRA ranks, $r \in \{8, 16, 32, 64\}$, for both the VE and LLM modules. We treat the projector as part of the vision encoder because it processes the same input stream for the same semantic purpose, so its adapter has the rank size of $r_{\text{ve}}$.
  
  \subsection{MARS Edge Case Handling} 
  We consider rare failure cases of the MARS rank search mechanism. Specifically, in very rare cases (0.72\% of all runs), the computed rank becomes complex-valued. This occurs when $E_{\text{ve}}$ is excessively large due to the inherent entropy of the data distribution. In such cases, we fall back to a default LoRA configuration by setting $r_{\text{ve}} = r_{\text{llm}}$.

  \section{Optimal VE Rank Size in the Small-Dataset Regime}
  \label{app:small-data-regime}
  
  \label{app:vlm_observations}
  
  \begin{figure}[h]
      \centering
      \includegraphics[width=1\linewidth]{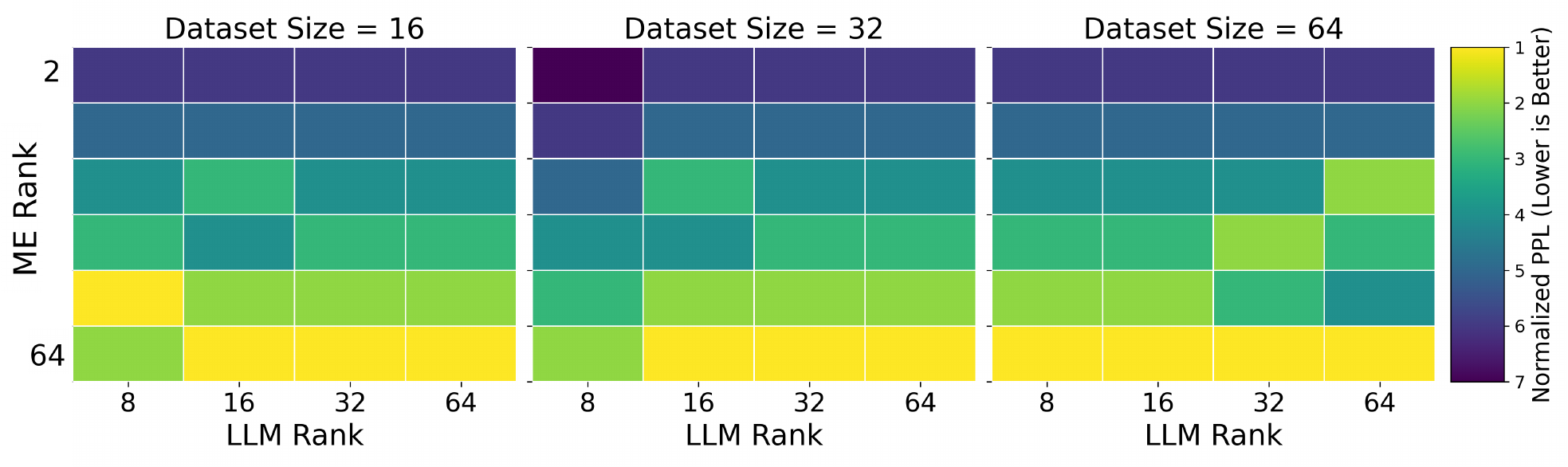}
      \caption{Optimal VE rank size in the small-dataset regime. Lower values (lighter colors) indicate better performance (i.e., lower perplexity).}
      \label{fig:smalldata_me}
  \end{figure}
  
  The optimal VE rank is strongly dependent on the dataset size, revealing a critical trade-off between rapid adaptation and overfitting. As illustrated in the heatmaps (Figure~\ref{fig:smalldata_me}), small-dataset regimes clearly prefer higher VE ranks. This is because the VE initially faces a large domain gap, and a higher rank provides the necessary adaptation capacity to quickly align with the LLM. From a capacity perspective, this aggressive approach allows the VE to absorb all available knowledge from limited samples, preventing it from becoming an early learning bottleneck.
  
  Conversely, in mid-to-large dataset regimes, a different pattern emerges: the preference shifts to smaller or more moderate VE ranks. The primary goal becomes reducing overfitting risk and ensuring the efficient utilization of model capacity. With a large and diverse set of examples, extreme adaptation capacity is no longer required. Instead of attempting to absorb all available knowledge, a moderate rank is sufficient to extract essential visual information without over-specializing on training examples. This provides the LLM with stable, moderately adapted inputs, which promotes balanced convergence, prevents overfitting, and allows the model to fully leverage the LLM's capacity for optimal modality harmonization.

  {\color{black}
  \section{Further Validation on Scaling Law-P}
  \label{app:detailed_sc_p}
  
  \paragraph{Formulation.} We formulate Scaling Law-P to model the fine-tuning loss (or perplexity, $\hat{L}$) for a VLM. This law extends existing frameworks by incorporating separate LoRA ranks for the vision encoder ($r_{\text{ve}}$) and the LLM ($r_{\text{llm}}$), alongside the dataset size ($D_f$):
  
  \begin{equation}
  \hat{L}(r_{\text{ve}}, r_{\text{llm}}, D_f) = A \cdot \frac{1}{(r_{\text{ve}})^{\alpha_m} \cdot (r_{\text{llm}})^{\alpha_l} \cdot D_f^{\beta}} + E
  \label{eq:scaling_law_p_appendix}
  \end{equation}
  
  ~\looseness=-1
  Here, $\alpha_m$ and $\alpha_l$ are scaling exponents for the VE and LLM ranks, reflecting their impact on loss reduction, while $\beta$ is the exponent for dataset size. $A$ and $E$ are fitted constants representing a scaling coefficient and an irreducible error, respectively. 
  
  Scaling Law-P is grounded in the foundational scaling laws proposed by prior work~\cite{zhang2024sc_llmft}, which were originally established in the context of unimodal LLM fine-tuning. Consequently, their scaling formulation resembles Equation~\ref{eq:scaling_law_p_appendix}, but treats $r_{ve}$ and $r_{llm}$ collectively as a single rank parameter $r$. \textbf{While this law holds under its original conditions, our work identifies a critical counterexample}: the law breaks down when there is a non-negligible imbalance in training dynamics across heterogeneous components (e.g., Vision Encoder \textit{vs.}\ LLM).
  
  \paragraph{Why Multiplicative (i.e., $r_{\mathrm{ve}} \cdot r_{\mathrm{llm}}$)?.}  We model the interaction as multiplicative because the VE and the LLM form a serial (not parallel) information-processing chain. A capacity bottleneck in the upstream encoder ($r_{\mathrm{ve}}$) imposes an irreversible loss of information that strictly bounds the maximum performance achievable by the downstream LLM, regardless of the LLM's capacity~\cite{ed2024gelora}.
  Also, following~\cite{zhang2024sc_llmft}, we compare the fitting error and find that a multiplicative formulation provides a better fit than an additive one.
  
      
  
  \subsection{Empirical Evidence as a Counterexample to the Original Scaling Laws}
  Figure~\ref{fig:appendix_sc1_val_pretrained} provides further empirical evidence of Scaling Law-P in pre-trained models.
  

  \begin{figure}[h]
    \centering
    \includegraphics[width=0.9\linewidth]{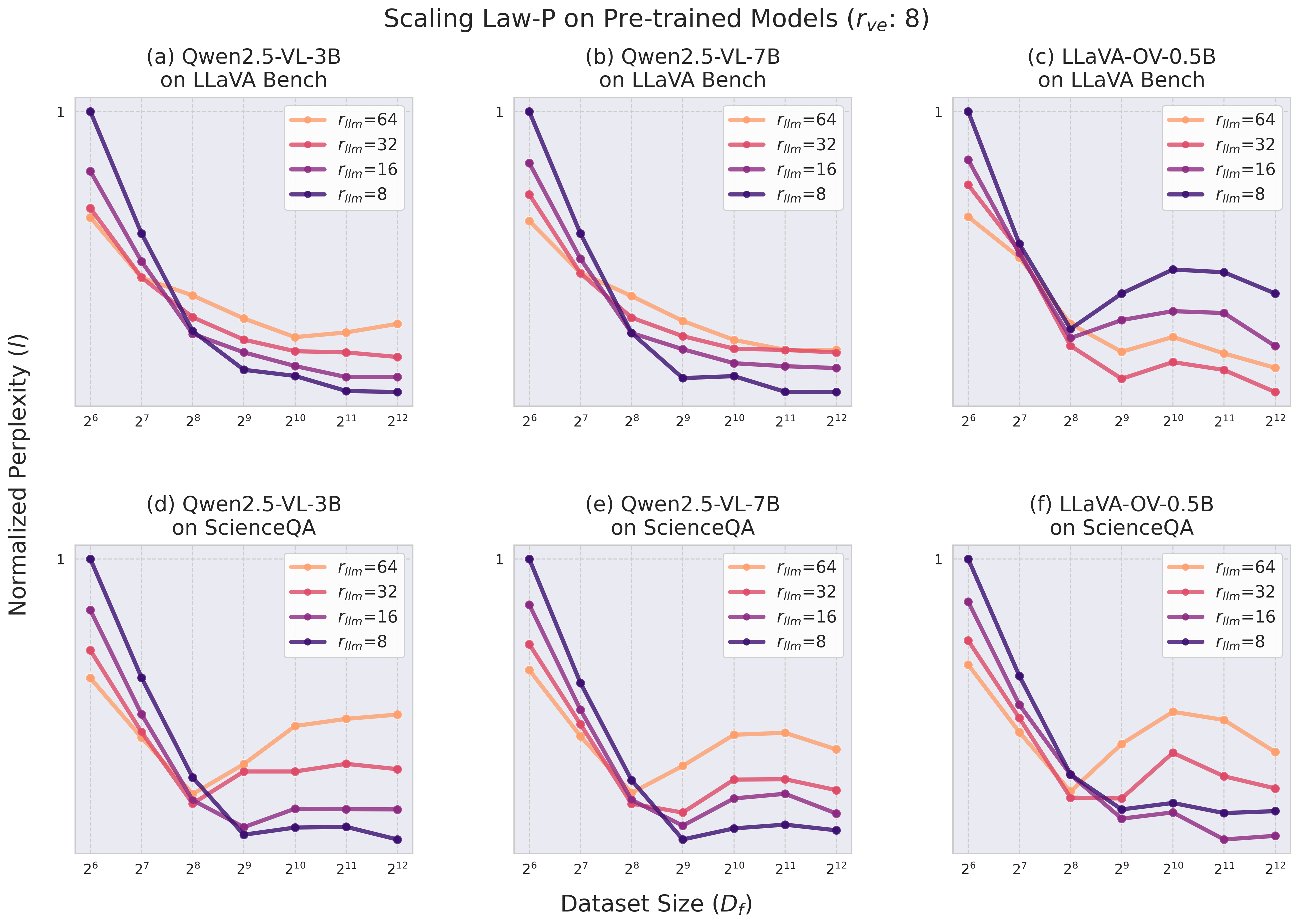}
    \caption{\textcolor{black}{Validating Scaling Law-P on Pre-trained Models}}
    \label{fig:appendix_sc1_val_pretrained}
  \end{figure}





  \section{Further Validation on Scaling Law-C: Theoretical and Empirical Generality}
  \label{app:detailed_sc_c}
  
  We propose Scaling Law-C to estimate the computational cost of convergence, $t$. We model $t$ as a tension between the \textbf{volume of information} to be absorbed ($D_f$) and the \textbf{condition number} of the optimization landscape ($r$).
  
  \paragraph{Formulation.}
  \label{sssec:scaling_law_2_formulation}
  Modeling the relationship between LoRA rank, dataset size, and the number of iterations required for convergence is a relatively underexplored area. To address this, we propose Scaling Law-C to model the convergence time for each modality-specific module within a VLM. For a given module $i$, where $i$ can represent a vision encoder (VE) or the LLM, we define the number of training steps to convergence, $t_i$, with the following general form:
  
  \begin{equation}
  t_{i}(r_{i}, D_f) = k_i \cdot (r_{i})^{\gamma_i} \cdot D_f^{\delta_i} + E_{i} 
  \label{eq:sc2_appendix}
  \end{equation}
  
  In this formulation, $t_i$ is the predicted training steps to convergence for module $i$ (e.g., $i \in \{\text{ve}, \text{llm}\}$). The LoRA rank for this module is $r_i$, and $D_f$ is the fine-tuning dataset size. The exponent $\gamma_i$ links the LoRA rank to convergence time; based on our observations that higher ranks reduce the number of steps, we fit this as a negative value. The exponent $\delta_i$ quantifies the impact of dataset size on convergence steps. Finally, $k_i$ and $E_i$ are constants fitted from experimental data for each module. This general formulation allows us to predict how the convergence time of any modality scales with its allocated LoRA rank and the amount of training data.
  
  \subsection{Theoretical Justification}
  \label{app:theoretical_justification}
  \begin{enumerate}
      \item \textbf{Information Absorption ($t \propto D_f^{\delta}$):} 
      Convergence time scales positively with dataset size ($\delta > 0$). We ground this in \textit{Knowledge Capacity Scaling Laws}~\cite{allen20232bit}, which establish that neural networks are bounded by a storage limit of approximately 2 bits of knowledge per parameter. As the dataset size $D_f$ increases, the total information content grows, pushing the fixed-rank adapter toward capacity saturation. To compress this increased entropy into the limited parameter budget, the optimizer requires a longer trajectory to resolve conflicting gradients and reach a generalization minimum, leading to the observed power-law increase in steps.

      \item \textbf{Geometric Conditioning ($t \propto r^{-\gamma}$):} 
      Convergence time scales inversely with rank ($\gamma > 0$). While \cite{mu2025lora_theory_converge} suggests that the theoretical worst-case convergence rate is rank-independent under ideal assumptions, our focus is more on the premise that rank fundamentally alters the \textit{geometry} of the loss landscape in practice.
     
      \begin{itemize}
      \item \textit{Stationary Point vs. Performance Threshold:} Previous analysis defines convergence as reaching \textit{any} stationary point; under this definition, the convergence rate is rank-independent \cite{mu2025lora_theory_converge}. However, Scaling Law-C models the time to reach a specific \textit{performance threshold}. Higher ranks are necessary to bridge the \textit{intrinsic dimension} gap, allowing the model to traverse toward the global optimum rather than stalling in rank-deficient local minima~\cite{ed2024gelora}.

      \item \textit{Spectral Stability:} \cite{shuttleworth2024intruderdimension} provides spectral evidence that low-rank constraints fundamentally alter the stability of weight updates. They find that low-rank adapters exhibit lower \textit{effective rank} (information density) compared to full fine-tuning. Increasing $r$ increases the effective rank of the update matrix, improving \textit{spectral stability} and ensuring smoother optimization dynamics that approximate full-rank behavior.

      \item \textit{Geometric Conditioning:} 
      Furthermore, higher ranks mitigate the formation of \textit{Intruder Dimensions}, which are high-magnitude singular vectors orthogonal to the pre-trained weights that appear in constrained low-rank regimes \cite{shuttleworth2024intruderdimension}. Low-rank models rely on these spectral anomalies, resulting in optimization instability and forgetting. Increasing the rank expands the optimization subspace, eliminating these artifacts and improving the condition number of the landscape. This accelerates the effective velocity of convergence ($t \downarrow$) by allowing the update trajectory to align with the smoother, pre-trained spectral manifold.

  \end{itemize}
  \end{enumerate}
  
  
  Theoretically, this parallelism indicates that the geometric conditioning induced by the rank ($r^{-\gamma}$) acts as an independent multiplicative preconditioner to the information complexity of the dataset ($D^{\delta}$). This supports our \emph{Subspace Optimization} hypothesis by decomposing the convergence time into two orthogonal factors.
  
  \subsection{Additional Empirical Evidence}  
  To demonstrate that this behavior reflects a fundamental property of low-rank adaptation, we investigate the Scaling Law-C pattern on the LLaVA-OneVision-0.5B and Qwen2.5-VL families (3B and 7B) using the ScienceQA dataset (Figures~\ref{fig:llava-ov-additional} and~\ref{fig: qwen_additional}). Consistent with the results observed in Figure~\ref{fig:sc2_b}, the Qwen2.5-VL models exhibit a near-linear relationship on the log--log scale, validating the power-law form $t \propto D^{\delta}$. Moreover, the fitted lines corresponding to different LoRA ranks remain nearly parallel across all architectures, providing strong empirical evidence for the separability of the rank and dataset size terms in the proposed scaling law (Equation~\ref{eq:sc2_appendix}). 
  
  \begin{figure}[h]
      \centering
      \begin{subfigure}[b]{0.48\textwidth}
          \centering
          \includegraphics[width=\textwidth]{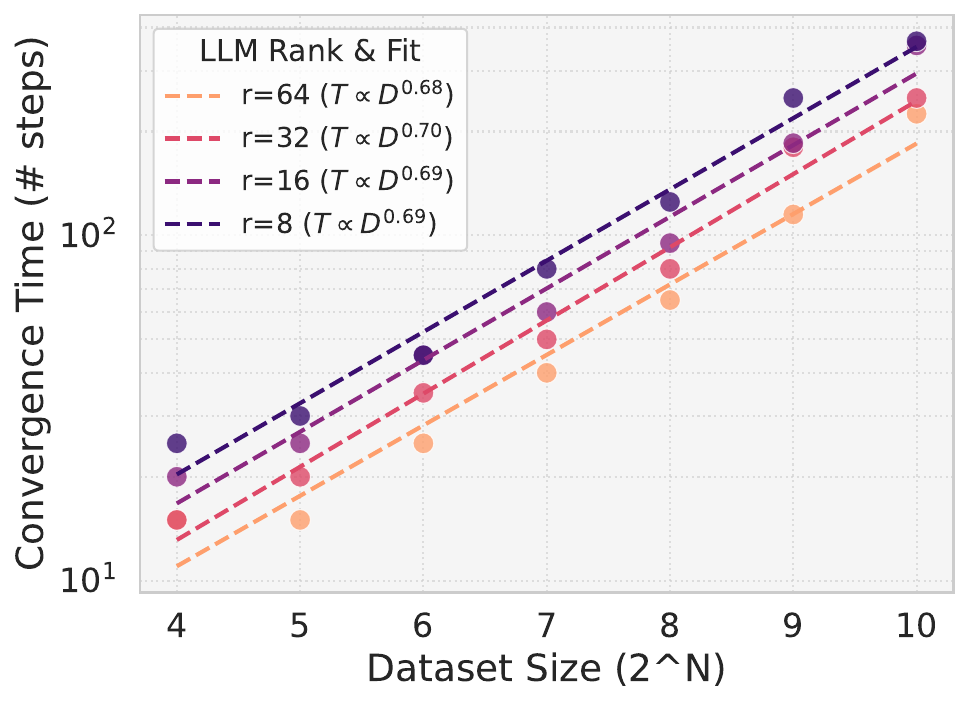} 
          \caption{LLM (LLaVA-OV-0.5B)}
          \label{fig:llm_scaling}
      \end{subfigure}
      \hfill
      \begin{subfigure}[b]{0.48\textwidth}
          \centering
          \includegraphics[width=\textwidth]{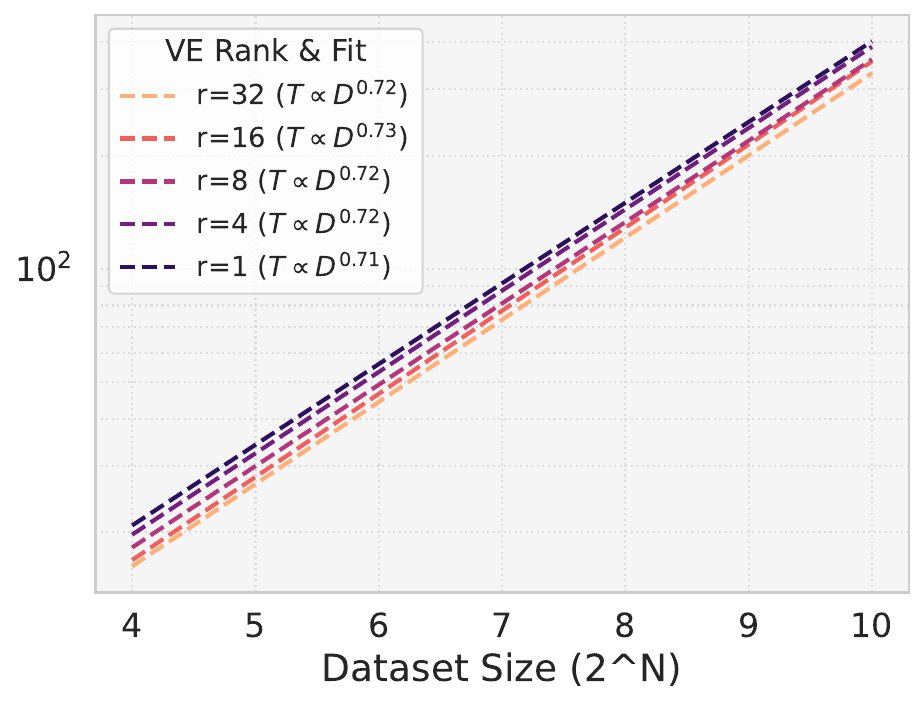}
          \caption{VE (LLaVA-OV-0.5B)}
          \label{fig:ve_scaling}
      \end{subfigure}
      \caption{Scaling Law-C Validation: LLaVA-OV-0.5B Models for the ScienceQA Dataset}
      \label{fig:llava-ov-additional}
  \end{figure}
  
  \begin{figure}[h]
      \centering
      \includegraphics[width=1\linewidth]{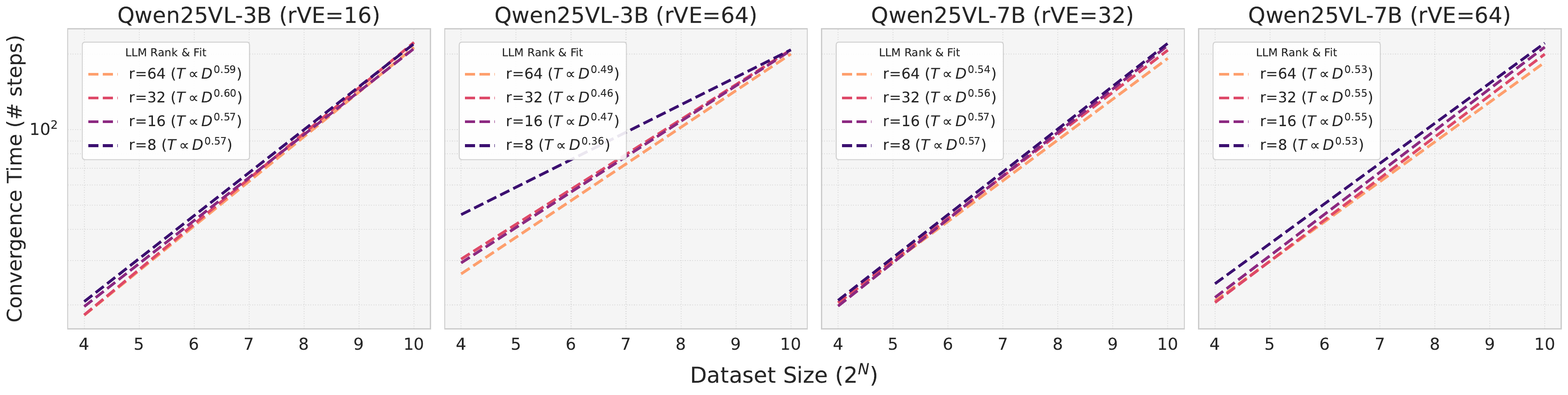}
      \caption{\textcolor{black}{Scaling Law-C Validation: Qwen2.5-VL Models for the ScienceQA Dataset}}
      \label{fig: qwen_additional}
  \end{figure}

  \newpage
  \section{Derivation of Equation~\ref{eq:r_ve_ideal}}
  \label{app:derivation}
  
  The objective of MARS is to minimize the gradient conflict caused by asynchronous convergence. By equating the convergence times derived from Scaling Law-C ($t_{ve} \approx t_{llm}$), we analytically solve for the optimal rank ratio. This ensures that both the visual perception and reasoning modules traverse their optimization manifolds at compatible velocities, preventing the fast module from overfitting while the slow module is still acquiring features.
  
  This section provides a step-by-step derivation of Equation~\ref{eq:r_ve_ideal}, which estimates the ideal VE-to-LLM parameter ratio ($r_{\text{ve}}$). The ideal ratio is defined as the value that balances the expected training time of the vision encoder ($t_{\text{ve}}$) and the LLM ($t_{\text{llm}}$) for a given target fine-tuning dataset size.
  
  The derivation starts with our two empirically fitted scaling law models for training time:
  
  \begin{align}
      t_{\text{ve}} &= k_v \cdot (r_{\text{ve}})^{\gamma_v} \cdot D_f^{\delta_v} + E_{\text{ve}} \\
      t_{\text{llm}} &= k_l \cdot (r_{\text{llm}})^{\gamma_l} \cdot D_f^{\delta_l} + E_{\text{llm}}
  \end{align}
  
  where $\Theta_{t\_{\text{ve}}} = \{k_v, \gamma_v, \delta_v, E_{\text{ve}}\}$ and $\Theta_{t\_{\text{llm}}} = \{k_l, \gamma_l, \delta_l, E_{\text{llm}}\}$ are the fitted coefficients. 
  
  To find the balance point, we set these two time estimates to be approximately equal ($t_{\text{ve}} \approx t_{\text{llm}}$) for a target dataset size, denoted as $D_{\text{target}}$. This gives the following relationship:
  
  \begin{equation}
  k_v \cdot (r_{\text{ve}})^{\gamma_v} \cdot D_{\text{target}}^{\delta_v} + E_{\text{ve}} \approx k_l \cdot (r_{\text{llm}})^{\gamma_l} \cdot D_{\text{target}}^{\delta_l} + E_{\text{llm}}
  \label{eq:r_ve_ideal_full2}
  \end{equation}
  
  Our goal is to solve for $r_{\text{ve}}$. We begin by isolating the term containing $r_{\text{ve}}$ on one side of the equation:
  
  \begin{equation}
  k_v \cdot (r_{\text{ve}})^{\gamma_v} \cdot D_{\text{target}}^{\delta_v} \approx k_l \cdot (r_{\text{llm}})^{\gamma_l} \cdot D_{\text{target}}^{\delta_l} + (E_{\text{llm}} - E_{\text{ve}})
  \end{equation}
  
  Finally, to solve for $r_{\text{ve}}$, we raise both sides to the power of $1/\gamma_v$, which yields the final expression\footnote{In very rare cases (0.72\% of our total runs) where the numerator becomes non-positive, implying that $E_{\text{ve}}$ is large due to the inherent entropy of the data distribution, we set $r_{\text{ve}} = r_{\text{llm}}$ to serve as a default LoRA configuration.}:
  
  \begin{equation}
  r_{\text{ve}} \approx \left( \frac{k_l \cdot (r_{\text{llm}})^{\gamma_l} \cdot D_{\text{target}}^{\delta_l} + E_{\text{llm}} - E_{\text{ve}}}{k_v \cdot D_{\text{target}}^{\delta_v}} \right)^{\frac{1}{\gamma_v}} 
  \label{eq:r_ve_ideal_full}
  \end{equation}

  \newpage
  \section{Evaluation on From-Scratch VLMs.}
  \label{app:from-scratch}

  To assess pure fine-tuning capability, we evaluate task performance on from-scratch models whose modules have not been exposed to multimodal training data. Since the original from-scratch models are not publicly available~\cite{li2024llava-ov, wang2024qwen2}, we instead assemble unimodal models and connect them with a projector to replicate the reference architectures (Sec.~\ref{sec:sc_setup}). As shown in Tables~\ref{tab:comparison_lr_combined_fs} and~\ref{tab:comparison_baselines_mars_fromscratch}, MARS consistently outperforms the baseline configurations, highlighting its effectiveness in enabling from-scratch models to acquire downstream knowledge.

  \begin{table}[h]
  \centering
  \footnotesize
  \begin{minipage}[t]{0.48\linewidth}
  \centering
  \caption{From-scratch: fixed-rank tuning across learning rates. Details as in Table~\ref{tab:comparison_lr_combined}.}
  \label{tab:comparison_lr_combined_fs}
  \resizebox{\linewidth}{!}{%
  \renewcommand{\arraystretch}{1.1}
  \begin{tabular}{l  l ccc c}
  \toprule
  \textbf{Model} & \textbf{Bench.} & \multicolumn{3}{c}{\textbf{LoRA} ($\textbf{lr}_{\textbf{ve}}$, $\textbf{lr}_{\textbf{llm}}$)}  & \textbf{MARS} \\
  \cmidrule(lr){3-5}
  &  & ($\bigstar$, 1e-5) & ($\bigstar$, 1e-6) & ($\bigstar$, 1e-7) & \\
  \midrule
  \multirow{2}{*}{OV-0.5B*}  & LLaVA ($\downarrow$) & \underline{5.0011} &	5.0209 &	5.4407 &	\textbf{4.9547} \\
  &SciQA ($\uparrow$) & \underline{42.49} & 42.12 & 39.88 & \textbf{45.01} \\
  
  \multirow{2}{*}{OV-7B*} & LLaVA ($\downarrow$) & \underline{3.8701}	& 4.0298 &	4.3407	& \textbf{3.5609} \\
  &SciQA ($\uparrow$)   & 63.71 & \underline{64.11} & 60.24 & \textbf{65.07} \\
  
  \multirow{2}{*}{Q2-VL-7B*}  & LLaVA ($\downarrow$) & \underline{3.7738}	& 3.7949	& 4.0315	& \textbf{3.6214} \\
  
  & SciQA ($\uparrow$) & \underline{65.47} & 62.87 & 59.28 & \textbf{66.47} \\
  
  \bottomrule
  \end{tabular}}
  \end{minipage}
  \hfill
  \begin{minipage}[t]{0.48\linewidth}
  \centering
  \caption{From-scratch: adaptive rank search baselines. Details as in Table~\ref{tab:comparison_baselines_mars}.}
  \label{tab:comparison_baselines_mars_fromscratch}
  \resizebox{\linewidth}{!}{%
  \renewcommand{\arraystretch}{1}
  \begin{tabular}{lccccc}
  \toprule
  \textbf{Model} & \multirow{2}{*}{\textbf{AdaLoRA}} & \multicolumn{3}{c}{\textbf{LoRA} ($\textbf{r}_{\textbf{ve}}$, $\textbf{r}_{\textbf{llm}}$)}  & \multirow{2}{*}{\textbf{MARS}} \\
  \cmidrule(lr){3-5}
  &   & ($\bigstar$, 8) & ($\bigstar$, 16) & ($\bigstar$, 32) &  \\
  \midrule
  \multicolumn{6}{l}{\textit{LLaVA Bench (perplexity $\downarrow$)} }  \\
  \midrule
  OV-0.5B* & 5.0807 & 5.2819   & 5.0011  & \underline{4.9912} & \textbf{4.9547} \\
  OV-7B* & 4.0012 & \underline{3.6175} & 3.8701 & 3.9204  & \textbf{3.5609} \\
  Q2-VL-7B* &  4.1555 & \underline{3.7480} & 3.7738 & 4.1057 & \textbf{3.6214}   \\
  
  \midrule
  \multicolumn{6}{l}{\textit{ScienceQA bench (accuracy (\%) $\uparrow$)}} \\
  \midrule
  OV-0.5B* &  41.52 & 38.27 & \underline{42.49} & 41.99  & \textbf{45.01} \\
  OV-7B* &  58.28   & 64.11 & \underline{64.15} & 64.07 & \textbf{65.07} \\
  Q2-VL-7B* & 60.08 & \underline{65.87} & 65.47 & 63.67  & \textbf{66.47} \\
  \bottomrule
  \end{tabular}}
  \end{minipage}
  \end{table}

  \section{Generality of MARS with Broader Benchmark Coverage}
  \label{appendix: generality_appendix_section}
  In Section~\ref{ssec:initial_mars_result}, we evaluate MARS on broader benchmark suites, including generalist suites (MME, POPE, MMStar) and domain-specific tasks (GQA, TextCaps, AI2D), as provided by the \textit{Multimodal Evaluation Toolkit}~\cite{zhang2024lmmeval1}. Here, we provide a brief description of each benchmark along with the detailed numerical results omitted from Figure~\ref{fig:generality}.
  
  
  \subsection{Benchmark Description}
  
  \begin{table}[h]
  \centering
  \caption{Summary of Benchmarks and Evaluation Metrics}
  \label{tab:benchmark_summary}
  \resizebox{0.8\linewidth}{!}{%
  \begin{tabular}{lp{5cm}p{5cm}l}
  \toprule
  \textbf{Benchmark} & \textbf{Focus} & \textbf{Metric in Figure~\ref{fig:generality}} \\ 
  \midrule
  MME~\cite{fu2023mme} & General perception and cognition & Cognition + Perception \\
  POPE~\cite{pope} & Object hallucination &  F1 Score \\
  MMStar~\cite{chen2024mmstar} &  Visual perception and reasoning & Avg. Accuracy \\
  GQA~\cite{hudson2018gqa} & Compositional visual reasoning &  Accuracy \\
  TextCaps~\cite{sidorov2019textcaps} & OCR-based image captioning & ROUGE\_L \\
  AI2D~\cite{ai2d}& Scientific diagram understanding & Accuracy \\
  \bottomrule
  \end{tabular}}
  \end{table}
  
  \subsection{Detailed Numbers for Figure~\ref{fig:generality}}
  \label{appendix: generality_appendix}
  \begin{table}[h]
  \centering
  \caption{Performance comparison on diverse multimodal benchmarks demonstrating the broad generalization of generalist models when MARS is applied to the connected VE and LLM.}
  \label{tab:performance_results}
  \resizebox{0.8\linewidth}{!}{%
  \begin{tabular}{lcccccc}
  \toprule
  \textbf{Method} & \textbf{MME (sum)} & \textbf{POPE} & \textbf{MMStar (avg)} & \textbf{TextCaps (Val)} & \textbf{AI2D} & \textbf{GQA} \\ 
  \midrule
  AdaLoRA~\cite{zhang2023adalora} & 1524.0 & 79.7 & 28.6 & 38.0 & 45.5 & 52.7 \\
  LoRA ($\bigstar$, 64)    & 1662.4 & 86.8 & 36.1 & 42.6 & 48.0 & 55.1 \\
  MARS    & 1695.6 & 86.3 & 39.4 & 42.9 & 49.9 & 55.7 \\
  \bottomrule
  \end{tabular}}
  \end{table}
  
  \begin{table}[h]
  \centering
  \renewcommand{\arraystretch}{1.2}
  \caption{Fine-grained capability breakdown on MMStar~\cite{chen2024mmstar}}
  \label{tab:cognitive_benchmarks}
  \resizebox{0.9\linewidth}{!}{%
  \begin{tabular}{lcccccc}
  \toprule
  \textbf{Method} & \textbf{Coarse Per.} & \textbf{Fine-Grained Per.} & \textbf{Instance Reason.} & \textbf{Logical Reason.} & \textbf{Sci \& Tech} & \textbf{Math} \\ 
  \midrule
  AdaLoRA & 34.7 & 23.2 & 27.8 & 29.9 & 24.4 & 31.5 \\
  LoRA ($\bigstar$, 64)     & 55.5 & 27.8 & 49.8 & 32.1 & 22.2 & 29.0 \\
  MARS    & 58.3 & 28.4 & 48.4 & 34.9 & 33.0 & 33.4 \\
  \bottomrule
  \end{tabular}%
  }
  \end{table}
    
\newpage
\section{Simultaneous Multi-Rank Tuning for Scalability}
\label{appendix:scalability}

\begin{figure}[h]
\centering
\includegraphics[width=0.4\linewidth]{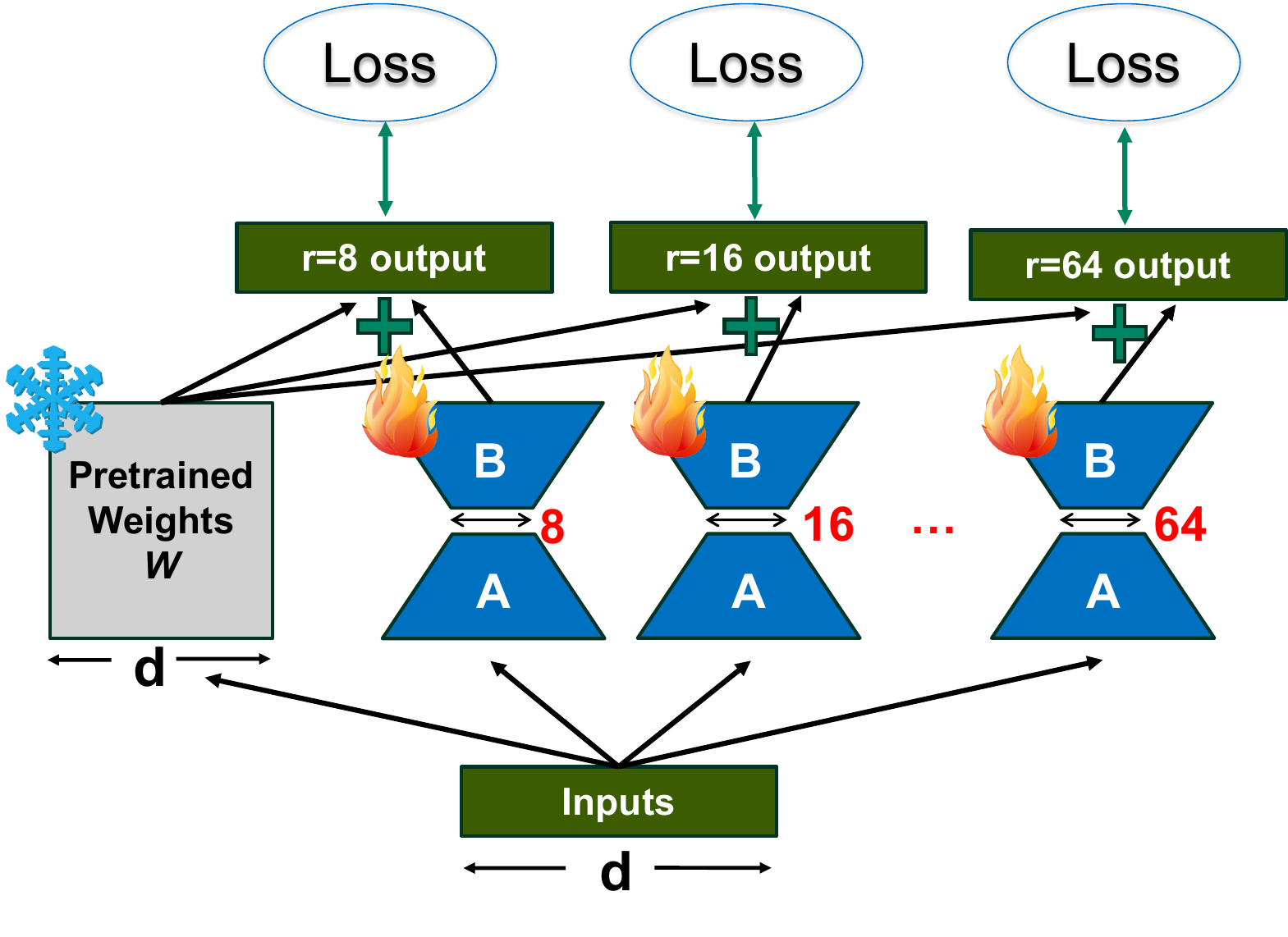}
\caption{Simultaneous Multi-Rank Tuning Mechanism}
\label{fig:multi-adpter-ft}
\end{figure}

Figure~\ref{fig:multi-adpter-ft} illustrates our proposed optimization for the calibration phase, leveraging the frozen nature of the pre-trained backbone weights ($W$). In standard fine-tuning, the computationally expensive forward pass through the large backbone ($W$) must be recomputed for every distinct rank configuration tested. To eliminate this redundancy, we construct a computational graph where the backbone features are computed only once per batch. These shared features are then branched into multiple, lightweight parallel LoRA adapter heads (e.g., initialized with ranks $r=\{8, 16, 32, 64\}$). Since the LoRA parameters constitute a negligible fraction of the total model size ($<1\%$), the computational overhead of running these parallel branches is minimal. This strategy allows us to gather perplexity and convergence statistics for all candidate ranks simultaneously in a single training run, in the best case, effectively reducing the calibration search cost from $\mathcal{O}(K^N)$ to $\mathcal{O}(1)$ (where $K$ is the number of rank candidates and $N$ is the number of modalities). This optimization ensures that MARS remains highly efficient even as the search space expands.

\section{Stability Across Runs}
\label{sec:stability}

We report mean$\pm$std over three seeds for MARS and the strongest LoRA baseline in Table~\ref{tab:stability}. Variance is low (std $< 0.3\%$) and rankings are stable across all settings.

\begin{table}[h]
\centering
\footnotesize
\caption{Stability of MARS across three random seeds. Performance is reported as mean$\pm$std.}
\label{tab:stability}
\begin{tabular}{llccc}
\toprule
\textbf{Model} & \textbf{Benchmark} & \textbf{Item} & \textbf{$(r_\text{ve}, r_\text{llm})$} & \textbf{Performance} \\
\midrule
LLaVA-OV-7B & LLaVABench & Baseline & (16, 16) & 2.2323$\pm$0.0038 \\
LLaVA-OV-7B & LLaVABench & MARS & (10, 16) & 2.1875$\pm$0.0025 \\
Qwen2.5-VL-3B & LLaVABench & Baseline & (16, 8) & 3.5914$\pm$0.0021 \\
Qwen2.5-VL-3B & LLaVABench & MARS & (10, 8) & 3.5825$\pm$0.0017 \\
\midrule
LLaVA-OV-7B & ScienceQA & Baseline & (16, 32) & 73.64$\pm$0.29\% \\
LLaVA-OV-7B & ScienceQA & MARS & (22, 32) & 74.30$\pm$0.25\% \\
Qwen2.5-VL-3B & ScienceQA & Baseline & (16, 8) & 78.32$\pm$0.18\% \\
Qwen2.5-VL-3B & ScienceQA & MARS & (15, 8) & 79.28$\pm$0.30\% \\
\bottomrule
\end{tabular}
\end{table}

\section{Transferability Analysis of Scaling Law Coefficients}
\label{sec:calibration_analysis}

To clarify the scope of transferability in MARS, we separately analyze the coefficients of Scaling Law-P and Scaling Law-C.

\paragraph{Scaling Law-P Coefficients.}
Table~\ref{tab:cv_slp} reports the coefficient of variation (CV) for each Scaling Law-P parameter. The dataset-size exponent $\beta$ and irreducible error $E$ transfer robustly across model families (CV $< 4\%$), indicating that asymptotic scaling behavior is architecture-agnostic. The rank exponents $\alpha_v$ and $\alpha_l$ are model-specific (cross-family CV $> 30\%$) and require per-model calibration.

\begin{table}[h]
\centering
\caption{Coefficient of Variation (CV) analysis for \textbf{Scaling Law-P} coefficients.}
\label{tab:cv_slp}
\resizebox{0.7\linewidth}{!}{%
\begin{tabular}{lccc}
\toprule
\textbf{Coefficient} & \textbf{Within-Family CV} & \textbf{Cross-Family CV} & \textbf{Transfers?} \\
\midrule
$\beta$ (dataset-size) & 2--3\% & 3.7\% & Yes \\
$E$ (irreducible loss) & $\sim$1\% & 1.1\% & Yes \\
$\alpha_v$ (VE rank) & 6--7\% & 31.1\% & No (need calibration) \\
$\alpha_l$ (LLM rank) & 4--6\% & 35.0\% & No (need calibration) \\
\bottomrule
\end{tabular}}
\end{table}

\paragraph{Scaling Law-C Coefficients.}
Table~\ref{tab:cv_slc} reports the CV for the convergence exponent $\delta$, which appears in Scaling Law-C (Equation~\ref{eq:scaling_law_2_general}). The high cross-family CV is precisely the justification for MARS's per-model calibration design.

\begin{table}[h]
\centering
\caption{Coefficient of Variation (CV) analysis for \textbf{Scaling Law-C} convergence exponent $\delta$.}
\label{tab:cv_slc}
\resizebox{0.5\linewidth}{!}{%
\begin{tabular}{lcc}
\toprule
\textbf{Grouping} & \textbf{Mean $\delta$} & \textbf{CV} \\
\midrule
Within LLaVA-OV (pre-trained) & 0.697 & 3.0\% \\
Within Qwen2.5-VL (pre-trained) & 0.570 & 2.5\% \\
Cross-family (all pre-trained) & 0.646 & 11.0\% \\
From-scratch vs.\ Pre-trained & 0.633 & 14.1\% \\
\bottomrule
\end{tabular}}
\end{table}

\section{Perplexity--Performance Correlation Across Benchmarks}
\label{app:ppl_correlation}

To validate that perplexity is a reliable optimization proxy beyond ScienceQA, we report Pearson correlations between validation perplexity and downstream task performance across six diverse benchmarks. As shown in Table~\ref{tab:ppl_correlation}, consistent negative correlations ($r$ ranging from $-0.72$ to $-0.97$) confirm that lower perplexity reliably predicts better task performance across both specialist and generalist benchmarks.

\begin{table}[h]
\centering
\caption{Correlation between validation perplexity and downstream task performance across diverse benchmarks.}
\label{tab:ppl_correlation}
\resizebox{0.5\linewidth}{!}{%
\begin{tabular}{llcc}
\toprule
\textbf{Benchmark} & \textbf{Fine-tuning Data} & \textbf{Pearson $r$} & \textbf{$p$-value} \\
\midrule
ScienceQA & ScienceQA & $-0.97$ & $< 0.001$ \\
MMStar & LLaVA-665K & $-0.89$ & $< 0.001$ \\
AI2D & LLaVA-665K & $-0.79$ & $< 0.01$ \\
MME & LLaVA-665K & $-0.78$ & $< 0.001$ \\
GQA & LLaVA-665K & $-0.74$ & $< 0.01$ \\
POPE & LLaVA-665K & $-0.72$ & $< 0.01$ \\
\bottomrule
\end{tabular}}
\end{table}

\section{Limitations}
\label{sec:limitations}
We acknowledge several limitations consistent with the empirical nature of prior scaling law research~\cite{kaplan2020sc_kaplan, zhang2024sc_llmft}:

\begin{enumerate}[leftmargin=1.5em]
    \item \textbf{Perplexity proxy scope.} While we demonstrate consistent negative correlations between perplexity and task performance across seven diverse benchmarks (Appendix~\ref{app:ppl_correlation}), the proxy has been validated primarily within our target evaluation suite. Its reliability on other task types (e.g., embodied AI, video understanding) requires further verification.
    \item \textbf{Multi-modality scalability (> 2 modalities).} Our tri-module experiment demonstrates $O(N)$ scalability,
    but exhaustive validation beyond two modalities (e.g., vision + audio + language~\cite{2026qwen-omni}) remains future work.
\end{enumerate}

\section{Broader Impacts}
\label{sec:impact}
As VLMs become more integrated into real-world applications, their adoption through fine-tuning is becoming increasingly popular across a wider range of downstream domains. 
The optimal rank pair identified by MARS reduces fine-tuning costs by eliminating the repetitive fine-tuning process caused by laborious trial-and-error hyperparameter searches. Ultimately, MARS will accelerate the development cycle and lower the carbon footprint for large-scale VLM fine-tuning.

\section{License of Assets}
\label{sec:lincese}
\noindent\textbf{Code and Frameworks.}
Our implementation is built on the Cornstarch framework\footnote{https://github.com/cornstarch-org/Cornstarch} (Apache-2.0).
For evaluation, we use LMMs-Eval\footnote{https://github.com/EvolvingLMMs-Lab/lmms-eval} (MIT).
We compare against AdaLoRA\footnote{https://github.com/QingruZhang/AdaLoRA} (MIT) and GeoLoRA\footnote{https://github.com/ScSteffen/Publication-GeoLoRA-Geometric-integration-for-parameter-efficient-fine-tuning} (MIT).

\noindent\textbf{Models.}
We use the following pre-trained models:
LLaVA-OneVision\footnote{https://github.com/LLaVA-VL/LLaVA-NeXT} (Apache-2.0),
and Qwen2-VL / Qwen2.5-VL\footnote{https://github.com/QwenLM/Qwen2-VL} (Apache-2.0).

\noindent\textbf{Datasets and Benchmarks.}
We utilize the following datasets and benchmarks:
LLaVA-Instruct\footnote{https://github.com/haotian-liu/LLaVA} (Apache-2.0),
ScienceQA\footnote{https://github.com/lupantech/ScienceQA} (MIT),
MME\footnote{https://github.com/BradyFU/Awesome-Multimodal-Large-Language-Models} (academic research use),
MMStar\footnote{https://github.com/MMStar-Benchmark/MMStar} (research use),
POPE\footnote{https://github.com/RUCAIBox/POPE} (MIT),
TextCaps\footnote{https://textvqa.org/textcaps/} (CC BY 4.0),
AI2D\footnote{https://prior.allenai.org/projects/diagram-understanding} (CC BY-SA),
and GQA\footnote{https://cs.stanford.edu/people/dorarad/gqa/} (CC BY 4.0).
All assets are used in accordance with their respective licenses and intended for research purposes.

  \newpage
  \section*{NeurIPS Paper Checklist}

\begin{enumerate}

\item {\bf Claims}
    \item[] Question: Do the main claims made in the abstract and introduction accurately reflect the paper's contributions and scope?
    \item[] Answer: \answerYes{} 
    \item[] Justification: Our core claim about imbalanced training dynamics reflects MARS’s contributions and scope.

\item {\bf Limitations}
    \item[] Question: Does the paper discuss the limitations of the work performed by the authors?
    \item[] Answer: \answerYes{} 
    \item[] Justification: Limitations are discussed in Appendix~\ref{sec:limitations}.

\item {\bf Theory assumptions and proofs}
    \item[] Question: For each theoretical result, does the paper provide the full set of assumptions and a complete (and correct) proof?
    \item[] Answer: \answerYes{} 
    \item[] Justification: See Appendices~\ref{app:derivation} for the proofs and~\ref{app:theoretical_justification} for the theoretical justification of our scaling laws. 

    \item {\bf Experimental result reproducibility}
    \item[] Question: Does the paper fully disclose all the information needed to reproduce the main experimental results of the paper to the extent that it affects the main claims and/or conclusions of the paper (regardless of whether the code and data are provided or not)?
    \item[] Answer: \answerYes{} 
    \item[] Justification: See Sec.\ref{sec:eval_setup} and Appendices\ref{app:exp_details} and~\ref{app:implementation_details}.

\item {\bf Open access to data and code}
    \item[] Question: Does the paper provide open access to the data and code, with sufficient instructions to faithfully reproduce the main experimental results, as described in supplemental material?
    \item[] Answer: \answerYes{} 
    \item[] Justification: We will release the source code and user guide in a personal GitHub repository.

\item {\bf Experimental setting/details}
    \item[] Question: Does the paper specify all the training and test details (e.g., data splits, hyperparameters, how they were chosen, type of optimizer) necessary to understand the results?
    \item[] Answer: \answerYes{} 
    \item[] Justification: All training and evaluation details are discussed in Appendices~\ref{app:dataset_setup} (in particular, the data splits) and~\ref{app:implementation_details} (in particular, the hyperparameters and training configurations).    

\item {\bf Experiment statistical significance}
    \item[] Question: Does the paper report error bars suitably and correctly defined or other appropriate information about the statistical significance of the experiments?
    \item[] Answer: \answerYes{} 
    \item[] Justification: See Appendix~\ref{sec:stability}. Due to the fine-tuning experiments and the large scope of evaluations, we performed statistical significance tests on representative cases only.

\item {\bf Experiments compute resources}
    \item[] Question: For each experiment, does the paper provide sufficient information on the computer resources (type of compute workers, memory, time of execution) needed to reproduce the experiments?
    \item[] Answer: \answerYes{} 
    \item[] Justification: We provide details on the computational resources in Sec.\ref{sec:eval_setup} (NVIDIA GH200 GPU) and Figure\ref{fig:gputime}.
    
\item {\bf Code of ethics}
    \item[] Question: Does the research conducted in the paper conform, in every respect, with the NeurIPS Code of Ethics \url{https://neurips.cc/public/EthicsGuidelines}?
    \item[] Answer: \answerYes{} 
    \item[] Justification: Our research conforms to the NeurIPS Code of Ethics.

\item {\bf Broader impacts}
    \item[] Question: Does the paper discuss both potential positive societal impacts and negative societal impacts of the work performed?
    \item[] Answer: \answerYes{} 
    \item[] Justification: See Appendix~\ref{sec:impact}.
    
\item {\bf Safeguards}
    \item[] Question: Does the paper describe safeguards that have been put in place for responsible release of data or models that have a high risk for misuse (e.g., pre-trained language models, image generators, or scraped datasets)?
    \item[] Answer: \answerNA{} 
    \item[] Justification: Our work does not introduce new data or models. Thus, MARS poses no such risks.

\item {\bf Licenses for existing assets}
    \item[] Question: Are the creators or original owners of assets (e.g., code, data, models), used in the paper, properly credited and are the license and terms of use explicitly mentioned and properly respected?
    \item[] Answer: \answerYes{} 
    \item[] Justification: See Appendix~\ref{sec:lincese}.

\item {\bf New assets}
    \item[] Question: Are new assets introduced in the paper well documented and is the documentation provided alongside the assets?
    \item[] Answer: \answerYes{} 
    \item[] Justification: We will release the source code and user guide in a personal GitHub repository (CC-BY 4.0 or Apache-2.0 license). 
    \begin{itemize}
    \item[] Guidelines: 
        \item The answer \answerNA{} means that the paper does not release new assets.
        \item Researchers should communicate the details of the dataset\slash code\slash model as part of their submissions via structured templates. This includes details about training, license, limitations, etc. 
        \item The paper should discuss whether and how consent was obtained from people whose asset is used.
        \item At submission time, remember to anonymize your assets (if applicable). You can either create an anonymized URL or include an anonymized zip file.
    \end{itemize}

\item {\bf Crowdsourcing and research with human subjects}
    \item[] Question: For crowdsourcing experiments and research with human subjects, does the paper include the full text of instructions given to participants and screenshots, if applicable, as well as details about compensation (if any)? 
    \item[] Answer: \answerNA{} 
    \item[] Justification: MARS does not involve crowdsourcing nor research with human subjects.

\item {\bf Institutional review board (IRB) approvals or equivalent for research with human subjects}
    \item[] Question: Does the paper describe potential risks incurred by study participants, whether such risks were disclosed to the subjects, and whether Institutional Review Board (IRB) approvals (or an equivalent approval/review based on the requirements of your country or institution) were obtained?
    \item[] Answer: \answerNA{} 
    \item[] Justification: MARS does not involve crowdsourcing nor research with human subjects.

\item {\bf Declaration of LLM usage}
    \item[] Question: Does the paper describe the usage of LLMs if it is an important, original, or non-standard component of the core methods in this research? Note that if the LLM is used only for writing, editing, or formatting purposes and does \emph{not} impact the core methodology, scientific rigor, or originality of the research, declaration is not required.
    \item[] Answer: \answerNA{} 
    \item[] Justification: The LLM is used only for writing, editing, literature review, coding, and formatting. The core methodological development in this research does not involve LLMs.

\end{enumerate}

  \end{document}